%% file: 00_main.tex
\documentclass{article} 
\usepackage{iclr2024_conference,times}

\input{math_commands.tex}

\usepackage{hyperref}
\usepackage{url}
\usepackage{graphicx}
\usepackage{booktabs}
\usepackage{multirow}
\usepackage{cleveref}
\usepackage{tikz}
\usepackage{amssymb}
\usepackage{pifont} 
\usepackage{wrapfig}
\usepackage{floatflt}
\usepackage{spverbatim}
\usepackage{xcolor,colortbl}
\usepackage{tcolorbox}
\usepackage{fancyvrb}
\usepackage{fvextra}

\newcommand{\cmark}{\ding{51}}%
\newcommand{\xmark}{\ding{55}}%

\title{Consistent Video-to-Video Transfer Using Synthetic Dataset}

\author{Jiaxin Cheng, Tianjun Xiao \& Tong He  \\
Amazon Web Services Shanghai AI Lab\\
\texttt{\{cjiaxin,tianjux,htong\}@amazon.com} \\
}

\newcommand{\ours}{InsV2V}

\iclrfinalcopy 
\newif\ifaddhead
\addheadfalse
\begin{document}

\maketitle

\input{00_abstract}
\input{01_intro}

\input{02_related}

\input{03_dataset}

\input{04_method}
\input{05_experiments}

\input{10_conclusion}

\bibliography{iclr2024_conference}
\bibliographystyle{iclr2024_conference}

\appendix
\include{11_appendix}

\end{document}

%% file: math_commands.tex
\usepackage{amsmath,amsfonts,bm}

\def\eqref#1{equation~\ref{#1}}

\def\1{\bm{1}}

\DeclareMathAlphabet{\mathsfit}{\encodingdefault}{\sfdefault}{m}{sl}
\SetMathAlphabet{\mathsfit}{bold}{\encodingdefault}{\sfdefault}{bx}{n}

\newcommand{\ie}{\textit{i.e.}}

%% file: 00_abstract.tex
\begin{abstract}
We introduce a novel and efficient approach for text-based video-to-video editing that eliminates the need for resource-intensive per-video-per-model finetuning. At the core of our approach is a synthetic paired video dataset tailored for video-to-video transfer tasks. Inspired by Instruct Pix2Pix's image transfer via editing instruction, we adapt this paradigm to the video domain. Extending the Prompt-to-Prompt to videos, we efficiently generate paired samples, each with an input video and its edited counterpart. Alongside this, we introduce the Long Video Sampling Correction during sampling, ensuring consistent long videos across batches. Our method surpasses current methods like Tune-A-Video, heralding substantial progress in text-based video-to-video editing and suggesting exciting avenues for further exploration and deployment. \url{https://github.com/amazon-science/instruct-video-to-video/tree/main}

\end{abstract}

\begin{figure}[!h]
    \centering
    \includegraphics[width=0.95\linewidth]{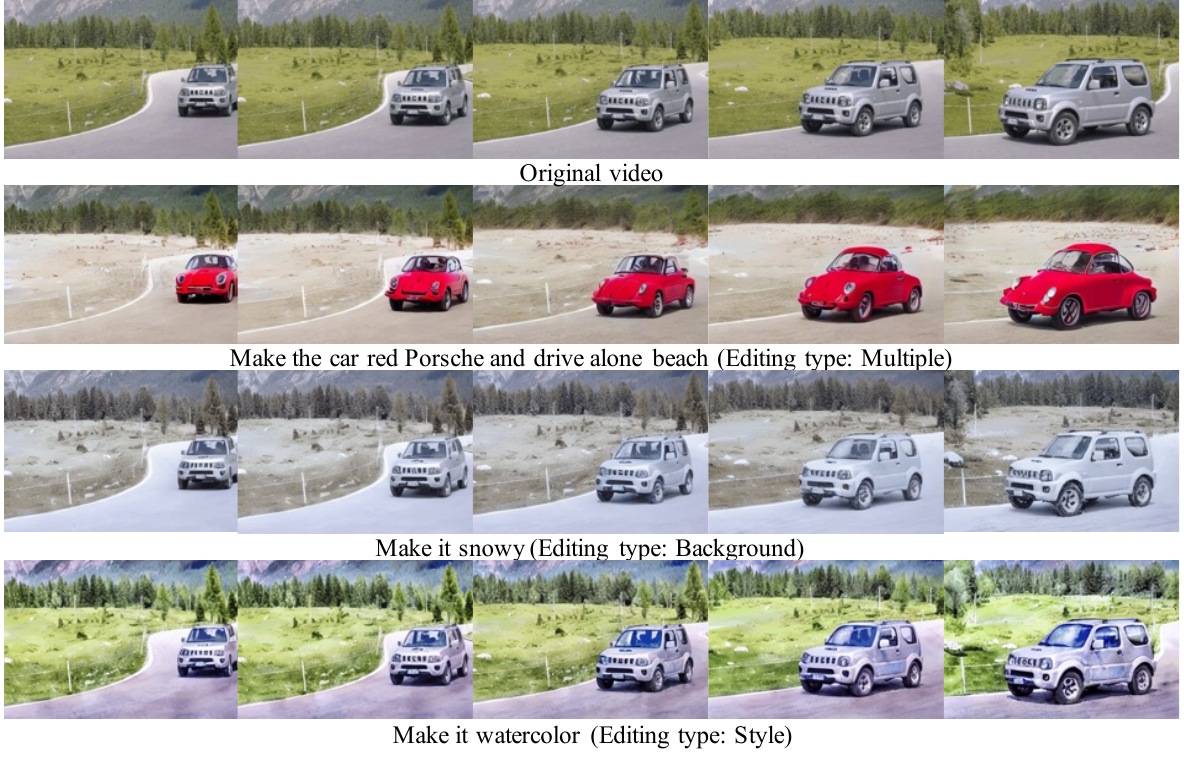}
    \caption{\ours\, has versatile editing capabilities encompassing background, object, and stylistic changing. Our method adopts a one-model-all-video strategy, achieving comparable performance while necessitating only inference. \ours\, eliminates the need to specify prompts for both original and target videos, simplifying the process by requiring only an edit prompt, thereby enhancing intuitiveness in video editing.}
    \label{fig:teaser}
\end{figure}

%% file: 01_intro.tex
\section{Introduction}
\label{sec:intro}


Text-based video editing~\cite{tuneavideo,zhao2023controlvideo,vid2vid,qi2023fatezero,liu2023video} has recently garnered significant interest as a versatile tool for multimedia content manipulation. However, existing approaches present several limitations that undermine their practical utility. Firstly, traditional methods typically require per-video-per-model finetuning, which imposes a considerable computational burden. Furthermore, current methods require users to describe both the original and the target video~\cite{tuneavideo,zhao2023controlvideo,vid2vid,qi2023fatezero,liu2023video}. This requirement is counterintuitive, as users generally only want to specify what edits they desire, rather than providing a comprehensive description of the original content. Moreover, these methods are constrained to individual video clips; if a video is too long to fit into model, these approaches fail to ensure transfer consistency across different clips.

To overcome these limitations, we introduce a novel method with several distinctive features. 
Firstly, our approach offers a universal one-model-all-video transfer, freeing the process from per-video-per-model finetuning. Moreover, our model simplifies user interaction by only necessitating an intuitive editing prompt, rather than detailed descriptions of both the original and target videos, to carry out desired alterations. Secondly, we develop a synthetic dataset precisely crafted for video-to-video transfer tasks. Through rigorous pairing of text and video components, we establish an ideal training foundation for our models. Lastly, we introduce a sampling method specifically tailored for generating longer videos. By using the transferred results from preceding batches as a reference, we achieve consistent transfers across extended video sequences. 

We introduce Instruct Video-to-Video (\ours), a diffusion-based model that enables video editing using only an editing instruction, eliminating the need for per-video-per-model tuning. This capability is inspired by Instruct Pix2Pix~\cite{brooks2023instructpix2pix}, which similarly allows for arbitrary image editing through textual instructions. A significant challenge in training such a model is the scarcity of naturally occurring paired video samples that can reflect an editing instruction. Such video pairs are virtually nonexistent in the wild, motivating us to create a synthetic dataset for training.

Our synthetic video generation pipeline builds upon a large language model (LLM) and the Prompt-to-Prompt~\cite{prompt2prompt} method that is initially designed for image editing tasks (\Cref{fig:synthetic_dataset}). We use an example-driven in-context learning approach to guide the LLM to produce these paired video descriptions. Additionally, we adapt the Prompt-to-Prompt (PTP) method to the video domain by substituting the image diffusion model with a video counterpart~\cite{ho2204video}. This modification enables the generation of paired samples that consist of an input video and its edited version, precisely reflecting the relationships delineated by the editing prompts.

In addressing the limitations of long video editing in conventional video editing methods, we introduce Long Video Sampling Correction (LVSC). This technique mitigates challenges arising from fixed frame limitations and ensures seamless transitions between separately processed batches of a lengthy video. LVSC employs the final frames of the previous batch as a reference to guide the generation of subsequent batches, thereby maintaining visual consistency across the entire video. We also tackle issues related to global or holistic camera motion by introducing a motion compensation feature that uses optical flow. Our empirical evaluations confirm the effectiveness of LVSC and motion compensation in enhancing video quality and consistency.

%% file: 02_related.tex
\section{Related Work}
\label{sec:related}

\textbf{Diffusion Models}
The advent of the diffusion model~\cite{sohl2015deep,ddpm} has spurred significant advancements in the field of image generation. Over the course of just a few years, we have observed the diffusion model making groundbreaking progress in a variety of fields. This includes areas such as super-resolution~\cite{saharia2022image}, colorization~\cite{saharia2022palette}, novel view synthesis~\cite{watson2022novel}, style transfer~\cite{zhang2023inversion}, and 3D generation~\cite{poole2022dreamfusion, tang2023make, cheng2023sdfusion}. These breakthroughs have been achieved through various means. Some are attributable to the enhancements in network structures such as Latent Diffusion Models (also known as Stable Diffusion)~\cite{ldm}, GLIDE~\cite{glide}, DALLE2~\cite{dalle2}, SDXL~\cite{podell2023sdxl}, and Imagen~\cite{imagen}. Others are a result of improvements made in the training paradigm~\cite{nichol2021improved,song2019generative,dhariwal2021diffusion,song2020score,ddim}. Furthermore, the ability to incorporate various conditions during image generation has played a crucial role. These conditions include elements such as layout~\cite{cheng2023layoutdiffuse,ldm}, segmentation~\cite{avrahami2023spatext,avrahami2022blended,balaji2022ediffi,yang2023paint}, or even the use of an image as reference~\cite{mou2023t2i,ruiz2023dreambooth,textualinversion}.

\textbf{Diffusion-based Text-Guided Image Editing} 
Image editing is a process where we don't desire completely unconstrained generation but modifying an image under certain guidance (\ie a reference image) during its generation. Various methods have been proposed to address this task. Simple zero-shot image-to-image translation methods, such as SDEdit~\cite{meng2021sdedit} performed through diffusion and denoising on reference image. Techniques that incorporate a degree of optimization, such as Imagic~\cite{kawar2023imagic}, which utilizes the concept of textual inversion~\cite{textualinversion}, and Null-text Inversion~\cite{mokady2023null}, which leverages the Prompt-to-Prompt strategy~\cite{prompt2prompt} to control the behavior of cross-attention in the diffusion model for editing, have also been explored. These methods can impede the speed of editing due to the necessity for per-image-per-optimization. Models like Instruct Pix2Pix~\cite{brooks2023instructpix2pix} have been employed to achieve image editing by training on synthetic data. This approach adeptly balances editing capabilities and fidelity to the reference image.

\textbf{Diffusion-based Text-Guided Video Editing} 
The success of the diffusion model in image generation has been extended to video generation as well~\cite{ho2022video,harvey2022flexible,blattmann2023align,mei2023vidm,ho2022imagen}, and similarly, text-guided video editing has sparked interest within the community. Techniques akin to those in image editing have found applications in video editing. For instance, Dreamix~\cite{molad2023dreamix} uses diffusion and denoising for video editing, reminiscent of the approach in SDEdit~\cite{meng2021sdedit}. Strategies altering the behavior of the cross-attention layer to achieve editing, like the Prompt-to-Prompt~\cite{prompt2prompt}, have been adopted by methods such as Vid2Vid-Zero~\cite{vid2vid}, FateZero~\cite{qi2023fatezero}, and Video-p2p~\cite{liu2023video}. Recent developments~\cite{wang2023videocomposer,gen1,zhao2023controlvideo} leverage certain condition modalities extracted from the original video, like depth maps or edge sketches, to condition the video generation. The related, albeit non-diffusion-based, method Text2live~\cite{bar2022text2live} also provides valuable perspectives on video editing.

%% file: 03_dataset.tex
\begin{figure}[th]
    \centering
    \includegraphics[width=\linewidth]{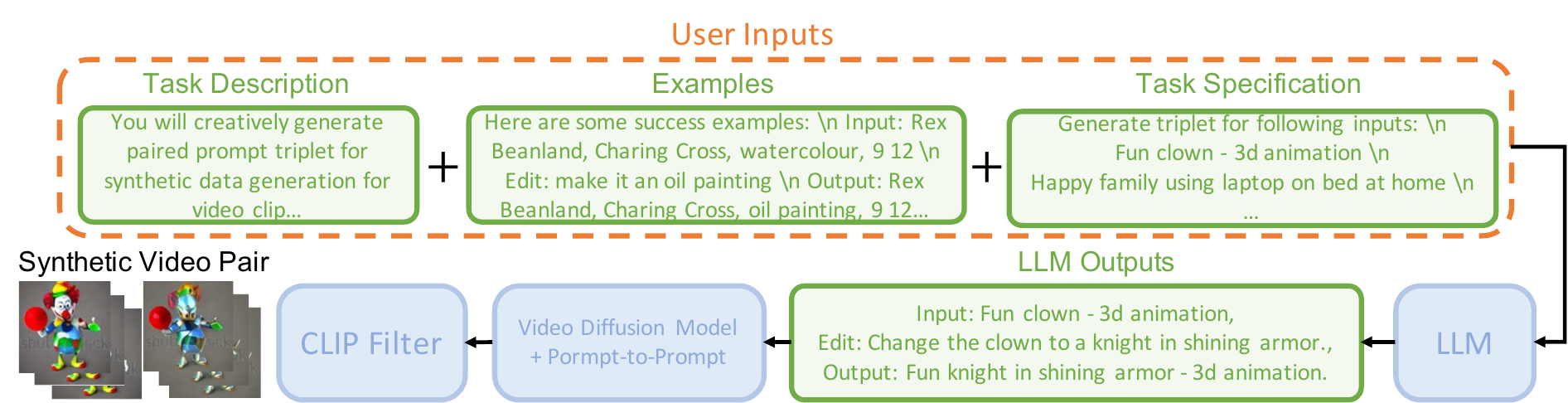}
    \caption{The pipeline for generating a synthetic dataset using a large language model, whose outputs include the prompt triplet consisting of input, edit, and edited prompts, as well as a corresponding pair of videos. Visualization of generated videos can be found in \Cref{sec.appendix_synthetic_vid}}
    \label{fig:synthetic_dataset}
\end{figure}

\section{Synthetic Paired Video Dataset}\label{sec.synthetic_data_pipe}



A crucial element for training a model capable of arbitrary video-to-video transfer, as opposed to a per-video-per-model approach, lies in the availability of ample paired training data. Each pair consists of an input video and its corresponding edited version, providing the model with a diverse range of examples essential for generalized performance. However, the scarcity of naturally occurring video pairs with such correspondence poses a significant challenge to the training process.

To address this, we introduce the concept of a trade-off between initial training costs, including dataset creation, and long-term efficiency. We advocate for the use of synthetic data, which, while incurring an upfront cost, accurately maintains the required correspondence and fulfills the conditions for effective video-to-video transfer learning. The merit of this synthetic data generation approach is underscored by its potential to offset the initial investment through substantial time savings and efficiency in the subsequent inference stages. This approach contrasts with `per-vid-per-model' methods that necessitate repetitive fine-tuning for each new video, making our strategy both cost-effective and practical in diverse real-world applications.

The potential of synthetic data generation, well-documented in the realm of text-guided editing~\cite{brooks2023instructpix2pix}, is thereby extended to video-to-video transfer. This method allows us to construct the optimal conditions for the model to learn, offering a practical solution to the inherent obstacles associated with acquiring matching real-world video pairs.

\subsection{Dataset Creation:}
In order to generate the synthetic dataset, we leverage the approach of Prompt-to-Prompt (PTP)~\cite{prompt2prompt}, a proven method for producing paired samples in the field of image editing. The PTP employs both self-attention and cross-attention replacements to generate semantically aligned edited images. In self-attention, the post-softmax probability matrix of the input prompt replaces that of the edited prompt. The cross-attention replacement specifically swaps the text embedding of the edited prompt with that of the input prompt. 

In the context of video-to-video transfer, we adapt PTP by substituting its underlying image diffusion models with a video diffusion model. In addition, we extend the self-attention replacement to temporal attention layers, a critical modification for maintaining structural coherence between input and edited videos. \Cref{fig:synthetic_dataset} shows the overall pipeline for data generation. To guide the synthetic data generation, we employ a set of paired text prompts, comprising an input prompt, an edited prompt, and an edit prompt. The input prompt corresponds to the synthetic original video, while the edited prompt and the edit prompt represent the desired synthetic edited video and the specific changes to be applied on the original video respectively. 

\subsection{Prompt Sources} 
Our synthetic dataset is constructed using paired prompts from two differentiated sources, each serving a specific purpose in the training process. The first source, LAION-IPTP, employs a fine-tuned GPT-3 model to generate prompts based on 700 manually labeled captions from the LAION-Aesthetics dataset~\cite{brooks2023instructpix2pix}. This yielded a set of 450,000 prompt pairs, of which 304,168 were utilized for synthetic video creation. While the GPT-3-based prompts offer a substantial volume of data, they originate from image captions and thus have limitations in their applicability to video generation. This led us to incorporate a second source, WebVid-MPT, which leverages video-specific captions from the WebVid 10M dataset~\cite{webvid}. Using MPT-30B~\cite{mpt} in a zero-shot manner, we devised a set of guidelines (see \Cref{sec:appendix_mpt} for details) to generate the needed paired prompts, adding an additional 100,000 samples to our dataset. Crucially, the WebVid-MPT source yielded a threefold increase in the success rate of generating usable samples compared to the LAION-IPTP source after sample filtering, reinforcing the need for video-specific captions in the training process. The LAION-IPTP prompt source demonstrated a success rate of 5.49\% (33,421 successful generations from 608,336 attempts). The WebVid-MPT prompt source showed a higher success rate of 17.49\% (34,989 successful generations from 200,000 attempts).

\subsection{Implementation Details and Sample Selection Criteria} 
\noindent\textbf{Implementation Details:} We use public available text-to-video model\footnote{https://modelscope.cn/models/damo/text-to-video-synthesis/summary} for generating synthetic videos. Each video has 16 frames, processed over 30 DDIM~\cite{ddim} steps by the diffusion model. The self-attention and cross-attention replacements in the Prompt-to-Prompt model terminate at a random step within the ranges of 0.3 to 0.45 and 0.6 to 0.85 respectively (out of 30 steps). The classifier-free guidance scale is a random integer value between 5 and 12.

\noindent\textbf{Sample Selection Criteria:} To ensure the quality of our synthetic dataset, we employ a CLIP-based filtering. For each prompt, generation is attempted with two random seeds, and three distinct clip scores are computed: CLIP Text Score: Evaluates the cosine similarity between each frame and the text.
CLIP Frame Score: Measures the similarity between original and edited frames.
CLIP Direction Score: Quantifies the similarity between the transition of original-frame-to-edited-frame and original-text-to-edited-text.
These scores are obtained for each of the 16 frames and averaged. A sample is preserved if it meets the following conditions: CLIP Text Score $>$ 0.2 for both original and edited videos, CLIP Direction Score $>$ 0.2, and CLIP Frame Score $>$ 0.5. Samples that fail to meet these criteria are discarded.


%% file: 04_method.tex
\begin{figure*}[t]
    \centering
    \includegraphics[width=\linewidth]{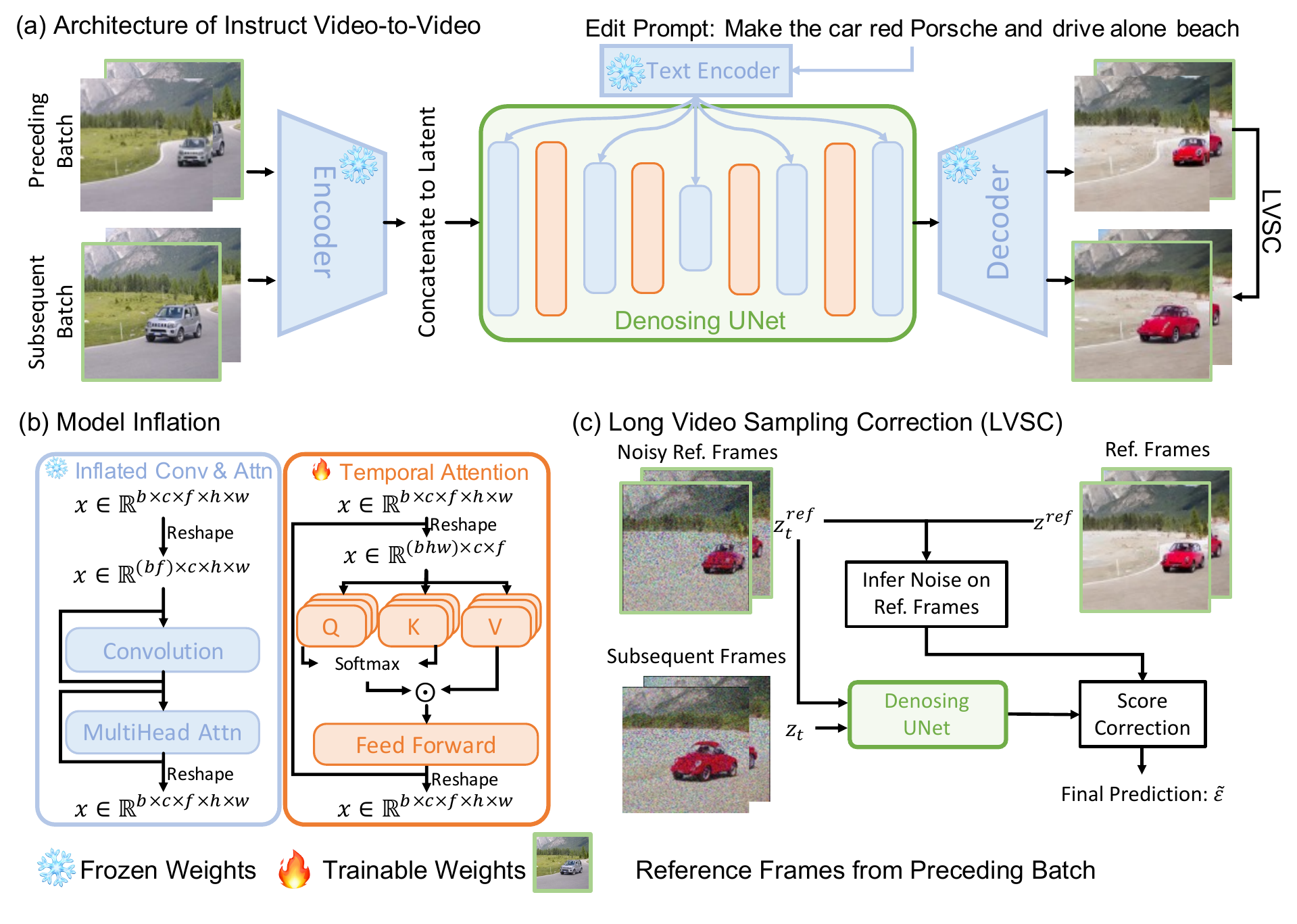}
    \caption{(a) The architecture of \ours. For handling long videos processed in multiple batches, our approach leverages the proposed LVSC to utilize the final frames from the preceding batch as reference frames for the subsequent batch. (b) The inflated convolutionals and attention layer, as well as temporal attention layer can handle 5D video tensors by dynamically reshaping them. (c) During each denoising iteration, the LVSC adjusts the predicted noise $\tilde{\varepsilon}_{\theta}(z_t)$ based on reference frames $z_t^{ref}$ prior to executing the DDIM denoising.}
    \label{fig:model}
\end{figure*}

\section{Model Architecture}
\label{sec:method}

\subsection{Preliminaries}

\textbf{Diffusion models} learn to predict content in an image by iteratively denoising an entirely random Gaussian noise. In the training process, the input image is first corrupted by the Gaussian noise, termed diffusion. The model's task is to restore a diffused noisy image to its original form. This process can be considered as the optimization of the variational lower bound on the distribution $p(x)$ of the image $x$ with a $T$ step Markovian inverse process. Various conditions $c$, such as text, images, and layouts, can be incorporated into diffusion models during the learning process. The model $\varepsilon_\theta$ we aim to train is conditioned on text and its training loss can be expressed as
\begin{equation}
    L = \mathbb{E}_{x,\varepsilon\sim\mathcal{N}(0,1),t}\|\varepsilon - \varepsilon_\theta(x_t, t, c)\|^2
\end{equation}

Where $x_t$ is the result of diffusing the input image $x$ at timestep $t \in [1, T]$ using random noise $\varepsilon$. In practice, we employ the Latent Diffusion Model (LDM)~\cite{ldm} as our backbone model and condition it on input videos by concatenating the videos to the latent space, as illustrated in \Cref{fig:model} (a). Instead of generating images in the RGB pixel space, LDM employs a trained Vector Quantized Variational AutoEncoder (VQVAE)~\cite{vqvae} to convert images to visual codes in latent space. This allows the model to achieve better results with the same training resources. Specifically, the image $x$ is first transformed into a latent code $z = \mathrm{VQEnc}(x)$, and the model learns to predict $p(z)$ from random noise. In the context of the video editing, two distinct conditions exist: the editing instructions and the reference video, represented by $c_T$ and $c_V$ respectively. The model is designed to predict $p(z)$ by optimizing the following loss function:

\begin{equation}\label{eq.ldm_objective}
    L = \mathbb{E}_{\mathrm{VQEnc}(x),\varepsilon\sim\mathcal{N}(0,1),t}\|\varepsilon - \varepsilon_\theta(z_t, t, c_V, c_T)\|^2
\end{equation}

\subsection{Inflate Image-to-Image Model To Video-to-Video Model}
Given the substantial similarities between image-to-image transfer and video-to-video transfer, our model utilizes a foundational pre-trained 2D image-to-image transfer diffusion model. Using foundational model simplifies training but falls short in generating consistent videos, causing noticeable jitter when sampling frames individually. Thus, we transform this image-focused model into a video-compatible one for consistent frame production. We adopt model inflation as recommended in previous studies~\cite{tuneavideo,guo2023animatediff}. This method modifies the single image diffusion model to produce videos. The model now accepts a 5D tensor input $x \in \mathbb{R}^{b \times c \times f \times h \times w}$. Given its architecture was designed for a 4D input, we adjust the convolutional and attention layers in the model (\Cref{fig:model} (b)). Our inflation process involves: (1) Adapting convolutional and attention layers to process a 5D tensor by reshaping it temporarily to 4D. Once processed, it's reverted to 5D. (2) Introducing temporal attention layers for frame consistency. When these layers handle a 5D tensor, they reshape it to a 3D format, enabling pixel information exchange between frames via attention.

\subsection{Sampling}
During sampling, we employ an extrapolation technique named Classifier-Free Guidance (CFG)~\cite{cfg} to augment the generation quality. Given that we have two conditions, namely the conditional video and the editing prompt, we utilize the CFG method for these two conditions as proposed in~\cite{brooks2023instructpix2pix}. Specifically, for each denoising timestep, three predictions are made under different conditions: the unconditional inference $\varepsilon_\theta(z_t,\varnothing,\varnothing)$, where both the conditions are an empty string and an all-zero video; the video-conditioned prediction $\varepsilon_\theta(z_t,c_V,\varnothing)$; and the video and prompt-conditioned prediction $\varepsilon_\theta(z_t,c_V,c_T)$. Here, we omit timestep $t$ for symbolic convenience. The final prediction is an extrapolation between these three predictions with video and text classifier-free guidance scale $s_V \geq 1$ and $s_T \geq 1$. 

\begin{align}
    \tilde{\varepsilon_\theta}(z_t, c_V, c_T) &= \varepsilon_\theta(z_t, \varnothing, \varnothing) \\ \nonumber
    &+ s_V \cdot (\varepsilon_\theta(z_t, c_V, \varnothing) - \varepsilon_\theta(z_t, \varnothing, \varnothing)) \\ \nonumber
    &+ s_T \cdot (\varepsilon_\theta(z_t, c_V, c_T) - \varepsilon_\theta(z_t, c_V, \varnothing))
\end{align}

\subsection{Long Video Sampling Correction for Editing Long Videos}\label{sec.long_vid_editing}


In video editing, models often face limitations in processing extended video lengths in one go. Altering the number of input frames can compromise the model's efficacy, as frame count is usually preset during training. To manage lengthy videos, we split them into smaller batches for independent sampling. While intra-batch frame consistency is preserved, inter-batch consistency isn't guaranteed, potentially resulting in visible discontinuities at batch transition points.

To address this issue, we propose Long Video Sampling Correction (LVSC): during sampling, the results from the final $N$ frames of the previous video batch can be used as a reference to guide the generation of the next batch (\Cref{fig:model} (c)). This technique helps to maintain visual consistency across different batches. Specifically, let $z^{ref} = z^{prev}_0[:, -N:] \in \mathbb{R}^{1, N, c, h, w}$ denote the last N frames from the transfer result of the previous batch. Here, to avoid confusion with the batch size, we set the batch size to 1. The ensuing batch is the concatenation of noisy reference frames and subsequent frames $[z^{ref}_t, z_t]$. On the model's prediction $\varepsilon_\theta(z_t) := \varepsilon_\theta(z_t, t, c_V, c_T)$, we implement a score correction and the final prediction is the summation between raw prediction $\varepsilon_\theta(z_t)$ and correction term $\varepsilon_t^{ref}-\varepsilon_\theta(z_t^{ref})$, where $\varepsilon_t^{ref}$ is the closed-form inferred noise on reference frames. For notation simplicity, we use $\varepsilon(z_t^{ref})$ and $\varepsilon(z_t)$ to denote the model's predictions on reference and subsequent frames, though they are processed together by the model instead of separate inputs.

\begin{align}
\varepsilon_t^{ref} &= \frac{z^{ref}_t - \sqrt{\Bar{\alpha_t}}z^{ref}}{\sqrt{1 - \Bar{\alpha_t}}} \in \mathbb{R}^{1,N,c,h,w}\label{eq.1} \\
\tilde{\varepsilon_\theta}(z_t) &= \varepsilon_\theta(z_t) + \frac{1}{N} \sum_{i=1}^N (\varepsilon_t^{ref}[:, i] - \varepsilon_\theta(z_t^{ref})[:, i])\label{eq.lvsc}
\end{align}

We apply averaging on the correction term when there are multiple reference frames as shown in \Cref{eq.1,eq.lvsc}, where $\Bar{\alpha_t}$ is the diffusion coefficient for timestep $t$ (\ie $z_t = \mathcal{N}(\sqrt{\Bar{\alpha_t}}z_0, (1 - \Bar{\alpha_t})I)$). In our empirical observations, we find that when the video has global or holistic camera motion, the score correction may struggle to produce consistent transfer results. To address this issue, we additionally introduce a motion compensation that leverages optical flow to establish correspondences between each pair of reference frames and the remaining frames in the batch. We then warp the score correction in accordance with this optical flow with details presented in~\Cref{sec.motion_compensation}.

%% file: 05_experiments.tex
\section{Experiments}\label{sec.expt}

\subsection{Experimental Setup}
\noindent\textbf{Dataset}
For our evaluation, we used the Text-Guided Video Editing (TGVE) competition dataset\footnote{https://sites.google.com/view/loveucvpr23/track4}. The TGVE dataset contains 76 videos that come from three different sources: Videov, Youtube, and DAVIS~\cite{DAVIS}. Every video in the dataset comes with one original prompt that describes the video and four prompts that suggest different edits for each video. Three editing prompts pertain to modifications in \textit{style}, \textit{background}, or \textit{object} within the video. Additionally, a \textit{multiple} editing prompt is provided that may incorporate aspects of all three types of edits simultaneously.


\noindent\textbf{Metrics for Evaluation}
Given that our focus is on text-based video editing, we look at three critical aspects. First, we assess whether the edited video accurately reflects the editing instructions. Second, we determine whether the edited video successfully preserves the overall structure of the original video. Finally, we consider the aesthetics of the edited video, ensuring it is free of imperfections such as jittering. Our evaluation is based on user study and automated scoring metrics. In the user study, we follow TGVE competition to ask users three key questions. The \textbf{Text Alignment} question: Which video better aligns with the provided caption? The \textbf{Structure} question: Which video better retains the structure of the input video? The \textbf{Quality} question: Aesthetically, which video is superior? These questions aim to evaluate the quality of video editing, focusing on the video's alignment with editing instructions, its preservation of the original structure, and its aesthetic integrity. For objective metrics, we incorporate PickScore~\cite{kirstain2023pick} that computes the average image-text alignment over all video frames and CLIP Frame (Frame Consistency)~\cite{CLIP}, which measures the average cosine similarity among CLIP image embeddings across all video frames. We prefer PickScore over the CLIP text-image score since it's tailored to more closely align with human perception of image quality, which is also noticed by~\cite{podell2023sdxl}.

\subsection{Baseline Methods}

We benchmark \ours\ against leading text-driven video editing techniques: Tune-A-Video~\cite{tuneavideo}, Vid2Vid-Zero~\cite{vid2vid}, Video-P2P~\cite{liu2023video}, and ControlVideo~\cite{zhao2023controlvideo}. Tune-A-Video has been treated as a \textit{de facto} baseline in this domain. Vid2Vid-Zero and Video-P2P adopt the cross-attention from Prompt-to-Prompt (PTP)\cite{prompt2prompt}, while ControlVideo leverages ControlNet\cite{controlnet}. We test all methods for 32 frames, but PTP-based ones, due to their computational demand, are limited to 8 frames. Baselines are processed in a single batch to avoid inter-batch inconsistencies and use latent inversion~\cite{mokady2023null} for structure preservation, which causes double inference time. Conversely, our method retains the video's structure more efficiently. 

We also extend the comparison to include recent tuning-free video editing methods such as TokenFlow~\cite{geyer2023tokenflow}, Render-A-Video\cite{yang2023rerender}, and Pix2Video\cite{ceylan2023pix2video}. These methods, by eliminating the necessity for individual video model tuning, present a comparable benchmark to our approach. To ensure frame-to-frame consistency, these methods either adopt cross-frame attention similar to Tune-A-Video\cite{tuneavideo}, as seen in~\cite{ceylan2023pix2video, yang2023rerender}, or establish pixel-level correspondences between features and a reference key frame as in~\cite{geyer2023tokenflow}. This approach is effective in maintaining quality when there are minor scene changes. However, in scenarios with significant differences between the key and reference frames, these methods may experience considerable degradation in video quality. This limitation is more clearly illustrated in \Cref{fig.additional_qua2} in the Appendix.

\subsection{Model Details}\label{sec.details}
Our model is adapted from a single image editing Stable Diffusion~\cite{brooks2023instructpix2pix} and we insert temporal attention modules after each spatial attention layers as suggested by~\cite{guo2023animatediff}. Our training procedure makes use of the Adam optimizer with a learning rate set at $5 \times 10^{-5}$. The model is trained with a batch size of 512 over a span of 2,000 iterations. This training process takes approximately 30 hours to complete on four NVIDIA A10G GPUs.

During sampling, we experiment with varying hyperparameters for video classifier-free guidance (VCFG) within the choice of [1.2, 1.5, 1.8], text classifier-free guidance to 10 and video resolutions of 256 and 384. A detailed visual comparison using these differing hyperparameters can be found in the supplementary material (\Cref{sec.pick_criteria}). The hyperparameters combination that achieves the highest PickScore is selected as the final sampling result. Each video is processed in three distinct batches using LVSC with a fixed frame count of 16 within a batch, including reference frames from preceeding batch, and resulting in a total frame count of 32.

\subsection{Long Video Score Correction and Motion Compensation}

\begin{wraptable}{l}{0.5\textwidth}
    \vspace{-0.7cm}
  \centering
  \small
  \caption{Comparison of motion-aware MSE and CLIP frame similarity between the last frame of the preceding batch and the first frame of the subsequent batch on TGVE dataset.}
  \begin{tabular}{c|c|c|c}
    \hline
    \textbf{LVSC} & \textbf{MC} &  \textbf{MAMSE (\%) $\downarrow$} & \textbf{CLIPFrame $\uparrow$} \\
    \cmidrule(r){1-1} \cmidrule(lr){2-2} \cmidrule(lr){3-3} \cmidrule(l){4-4}
    \xmark & \xmark & 2.02 & 0.9072 \\
    \cmark & \xmark & 1.44 & 0.9093 \\
    \cmark & \cmark & 1.37 & 0.9095 \\
    \hline
  \end{tabular}\label{tab.lvsc}
    \vspace{-0.2cm}
\end{wraptable}

To assess performance improvement, we employ CLIP Frame Similarity and Motion-Aware Mean Squared Error (MAMSE) as evaluation metrics. Unlike traditional MSE, MAMSE accounts for frame-to-frame motion by utilizing optical flow to warp images, thereby ensuring loss computation in corresponding regions. Incorporating Long Video Score Correction (LVSC) and Motion Compensation (MC) has led to enhanced performance as reflected in~\Cref{tab.lvsc}. Further qualitative comparison, detailing the benefits of LVSC and MC, are provided in ~\Cref{sec.long_vid_editing_appendix,sec.motion_compensation}.



    

\subsection{User Study}\label{sec.user_study}



\begin{table}[h]
\centering
\small
\caption{The first two columns display automated metrics concerning CLIP frame consistency and PickScore. The final four columns pertain to a user study conducted under the TGVE protocol, where users were asked to select their preferred video when comparing the method against the TAV.}
\begin{tabular}{lcccccc}
\hline
\textbf{Method} & \textbf{CLIPFrame} & \textbf{PickScore} & \textbf{Text Alignment} & \textbf{Structure} & \textbf{Quality} & \textbf{Average} \\
\cmidrule(r){1-1} \cmidrule(lr){2-2} \cmidrule(lr){3-3} \cmidrule(lr){4-4} \cmidrule(lr){5-5} \cmidrule(lr){6-6} \cmidrule(l){7-7}
TAV* & 0.924 & 20.36 & - & - & - & - \\
CAMP* & 0.899 & 20.71 & 0.689 & 0.486 & 0.599 & 0.591 \\
T2I\_HERO* & 0.923 & 20.22 & 0.531 & 0.601 & 0.564 & 0.565 \\
\cmidrule(r){1-1} \cmidrule(lr){2-2} \cmidrule(lr){3-3} \cmidrule(lr){4-4} \cmidrule(lr){5-5} \cmidrule(lr){6-6} \cmidrule(l){7-7}
Vid2Vid-Zero& 0.926 & 20.35 & 0.400 & 0.357 & 0.560 & 0.439 \\
Video-P2P & 0.935 & 20.08 & 0.355 & 0.534 & 0.536 & 0.475 \\
ControlVideo & 0.930 & 20.06 & 0.328 & 0.557 & 0.560 & 0.482 \\
\cmidrule(r){1-1} \cmidrule(lr){2-2} \cmidrule(lr){3-3} \cmidrule(lr){4-4} \cmidrule(lr){5-5} \cmidrule(lr){6-6} \cmidrule(l){7-7}
TokenFlow & \textbf{0.940} & 20.49 & 0.287 & 0.563 & 0.624 & 0.491 \\
Pix2Video & 0.916 & 20.12 & 0.468 & 0.529 & 0.538 & 0.511 \\
Render-A-Video & 0.909 & 19.58 & 0.326 & 0.551 & 0.525 & 0.467 \\
\cmidrule(r){1-1} \cmidrule(lr){2-2} \cmidrule(lr){3-3} \cmidrule(lr){4-4} \cmidrule(lr){5-5} \cmidrule(lr){6-6} \cmidrule(l){7-7}
\ours\,(Ours)  & 0.911 & \textbf{20.76} & \textbf{0.690} & \textbf{0.717} & \textbf{0.689} & \textbf{0.699}\\
\hline
\end{tabular}\\
\raggedright
\hspace{10px} \scriptsize{*: Scores from \href{https://huggingface.co/spaces/loveu-tgve/loveu-tgve-leaderboard}{TGVE leaderboard}.}
\label{tab:protocol1}
\end{table}

\begin{figure*}[!h]
    \centering
    \includegraphics[width=\linewidth]{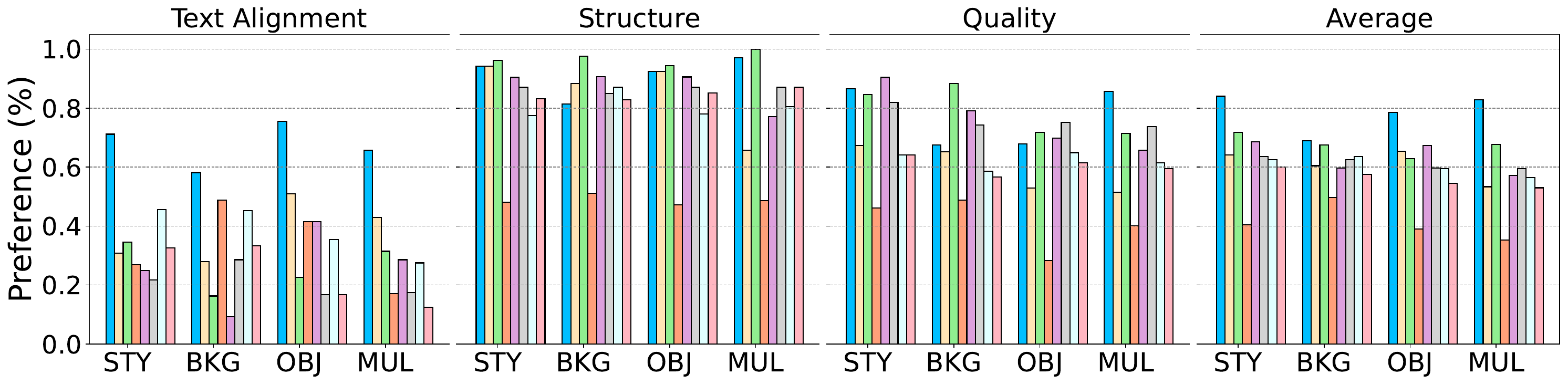}
    \includegraphics[width=\linewidth]{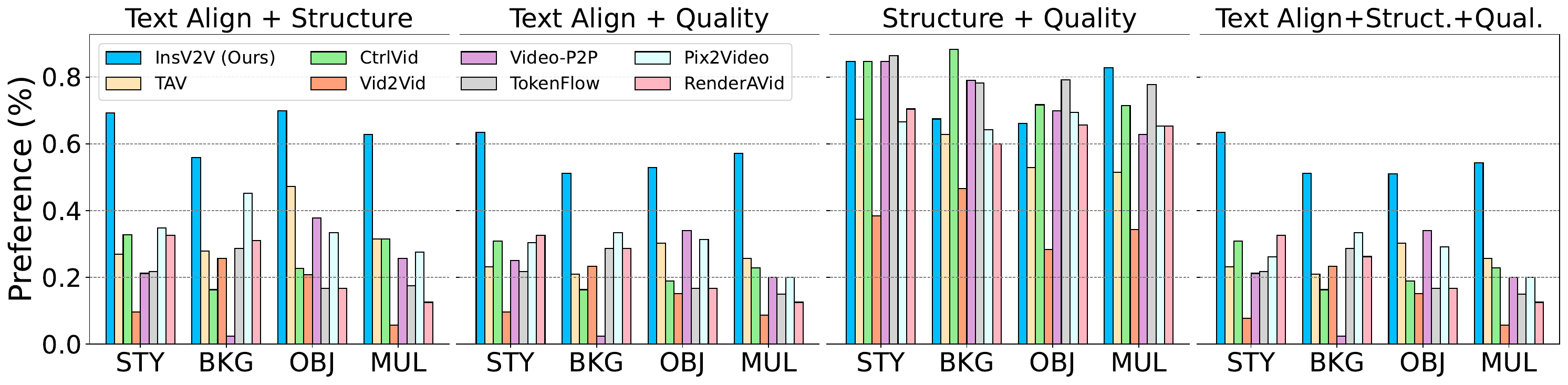}
    \caption{The abbreviations on x-axis indicate user preferences across four types of video editing within TGVE: Style, Background, Object, and Multiple. Each title specifies the evaluation metrics used for the corresponding figures. A "+" symbol signifies that the user vote meets multiple criteri. Additional qualitative results are presented in \Cref{sec.qualitative_comparison}
    }
    \label{fig:protocol2}
\end{figure*}

We conducted two separate user studies. The first followed the TGVE protocol, where we used Tune-A-Video as the baseline and compared the outcomes of our method with this well-known approach. However, we realized that this method of always asking users to choose a better option might not fully represent the reality, where both videos could either perform well or poorly. Thus, in the second user study, we compare our method with seven publicly available baselines. Instead of asking users to choose a better video, we asked them to vote for the quality of text alignment, structural preservation, and aesthetic quality for each transferred video. As evidenced by \Cref{tab:protocol1} and \Cref{fig:protocol2}, our approach excelled in all metrics except CLIP Frame similarity. We contend that CLIP Frame similarity is not an entirely apt metric because it only captures semantic similarity between individual frames, which may not be constant throughout a well-edited video due to changing scenes.

In our evaluation protocol, we additionally introduce a multi-metric assessment, capturing cases where videos satisfy multiple evaluation criteria concurrently. This composite measure addresses a key shortcoming of single metrics, which may inadequately reflect overall editing performance. For example, a high structure score might indicate that the edited video perfectly preserves the original, but such preservation could come at the expense of alignment with the editing instruction.

The results presented in \Cref{fig:protocol2} further corroborate the advantages of our approach. While baseline methods demonstrate respectable performance when text alignment is not a prioritized criterion, they fall short when this element is incorporated into the assessment. In contrast, our method not only excels in aligning the edited video with the textual instructions but also maintains the structural integrity of the video, resulting in a high-quality output. This dual success underlines the robustness of our method in meeting the nuanced requirements of video editing tasks.




%% file: 10_conclusion.tex
\section{Conclusion}
\label{sec:conclusion}

This research tackles the problem of text-based video editing. We have proposed a synthetic video generation pipeline, capable of producing paired data required for video editing. To adapt to the requirements of video editing, we have employed the model inflation technique to transform a single image diffusion model into a video diffusion model. As an innovation, we've introduced the Long Video Sampling Correction to ensure the generation of consistent long videos. Our approach is not only efficient but also highly effective. The user study conducted to evaluate the performance of our model yielded very high scores, substantiating the robustness and usability of our method in real-world applications.

%% file: 11_appendix.tex
\appendix
\section*{\centering Supplementary of Coherent Video-to-Video Transfer Using Synthetic Datasets}
\section{Instruction For Zero-shot MPT Data Generation:}\label{sec:appendix_mpt}

The instruction for MPT comprises three parts. The first is the task description which delineates the objectives for the bot.

\begin{Verbatim}[breaklines=true]
You are a bot to generate synthetic text data for generating video clip. You will creatively generate paired prompt triplet for synthetic data generation for video clip. You will be given an input prompt and you should return the edit prompt and output prompt. The output prompt reflects the sentence after applying edit prompt on the input prompt. Ensure that the prompt are proper for generating video clip (i.e. prompt should describe a scene).

Successful editing do changing the main subject, modifying the context or setting or altering the artistic style. 

Here are some examples of editing that are likely to success (do not limit to the verb used in the follow. You must be creative and the editing should be diverse):

Edit of Landscape:
Edit: Convert the cityscape to a seascape.
Edit: Turn the desert scene into a lush forest.

Replacement of Characters:
Edit: Replace the cowboys with astronauts.
Edit: Turn the group of children into a group of elderly people.

Edit of Time:
Edit: Switch the night scene to a day scene.
Edit: Transform the contemporary setting into a medieval setting.

Addition of Significant Elements:
Edit: Add a full moon to the clear sky.
Edit: Include a rainbow in the cloudy scene.

Edit of Weather or Season:
Edit: Make the sunny day into a snowfall.
Edit: Transform the summer scene into autumn.

Edit the Action or Activity:
Edit: Change the soccer game to a ballet performance.
Edit: Replace the cooking scene with a gardening scene.

Edit of Artistic Style:
Edit: Make it look like a watercolor painting.
Edit: It is now in the style of Van Gogh. (do not only use Van Gogh)

\end{Verbatim}

The subsequent phase of the instruction involves presenting MPT with five randomly selected examples from the LAION-IPTP dataset. These samples have been previously successful in generating paired prompts that meet the CLIP filter criteria. The depiction below illustrates this process:

\begin{Verbatim}[breaklines=true]
Here are some success examples (please be creative and not limited to examples)

Input: Graham Wands - George Square, Glasgow, watercolour
Edit: Turn the watercolour into a pencil sketch
Output: Graham Wands - George Square, Glasgow, pencil sketch

Input: Pierre de Clausade, (French, 1910-1976), Winter at the Lake
Edit: make it a sunset
Output: Pierre de Clausade, (French, 1910-1976), Sunset at the Lake

Input: Rex Beanland, Charing Cross, watercolour, 9 12
Edit: make it an oil painting
Output: Rex Beanland, Charing Cross, oil painting, 9 12

Input: Mark Van Crombrugge, Old Milk Bottle and Grapes, oil, 31 x 59.
Edit: Make the bottle transparent.
Output: Mark Van Crombrugge, Transparent Old Milk Bottle, oil, 31 x 59.

Input: """Large Original Oil painting on canvas. Beautiful portrait of a woman 24x24"""""""
Edit: make the woman a cat
Output: """Large Original Oil painting on canvas. Beautiful portrait of a cat 24x24""""

\end{Verbatim}

Finally, MPT is provided with video captions from the WebVid dataset that are designated for processing. This marks the initiation of the generation process for novel, paired prompts.

\begin{Verbatim}[breaklines=true]
Generate triplet for following inputs:

Merida, mexico - may 23, 2017: tourists are walking on a roadside near catholic church in the street of mexico at sunny summer day.

Fun clown - 3d animation

Happy family using laptop on bed at home

11th march 2017. nakhon pathom, thailand. devotees goes into a trance at the wai khru ceremony at wat bang phra temple. what bang phra is famous for its magically charged tattoos and amulets.

Decorate with pineapple sweet cake roll.

Beautiful lake aerial view

Frankfurt, germany-circa 2013:traffic with skyscrapers in background at night along the river main in frankfurt, time lapse

Young positive couple laughing in the backyard in front of the large house under falling snow. bearded man and attractive woman in warm clothes have winter fun. the guy showing thumb up

Broadcast twinkling squared diamonds, multi color, abstract, loopable, 4k

Wheat harvesting. combine harvester gathers the wheat crop on the field.
\end{Verbatim}

\section{Visualization of Synthetic Video Dataset}\label{sec.appendix_synthetic_vid}
\begin{figure}
    \centering
    \includegraphics[width=\linewidth]{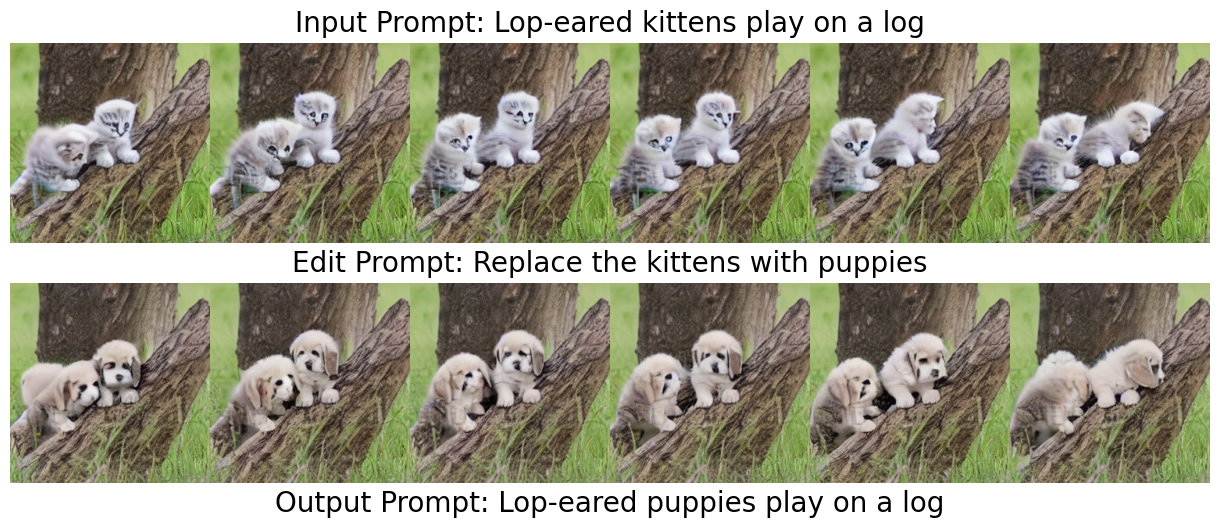}
    \includegraphics[width=\linewidth]{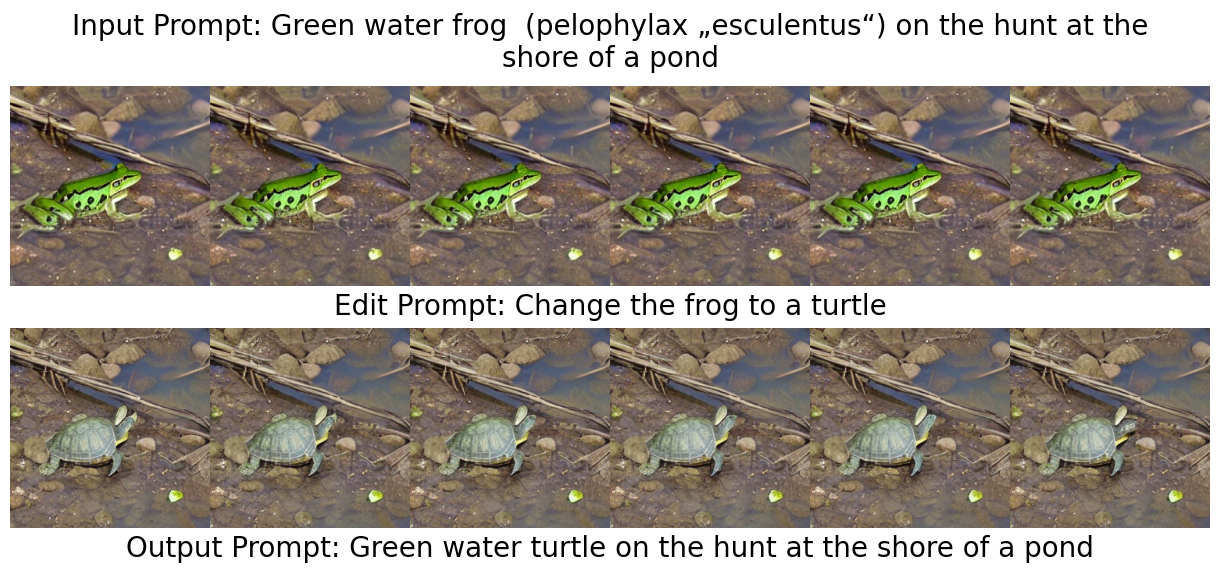}
    \caption{Visualization of synthetic samples produced by the data generation pipeline in \Cref{sec.synthetic_data_pipe}. Each sample contains two generated videos: the upper one represents the input prompt, while the lower one depicts the output prompt. The actual videos comprise 16 frames while 6 subsampled frames are displayed here for brevity.}
    \label{fig.synthetic_vid1}
\end{figure}

\begin{figure}
    \centering
    \includegraphics[width=\linewidth]{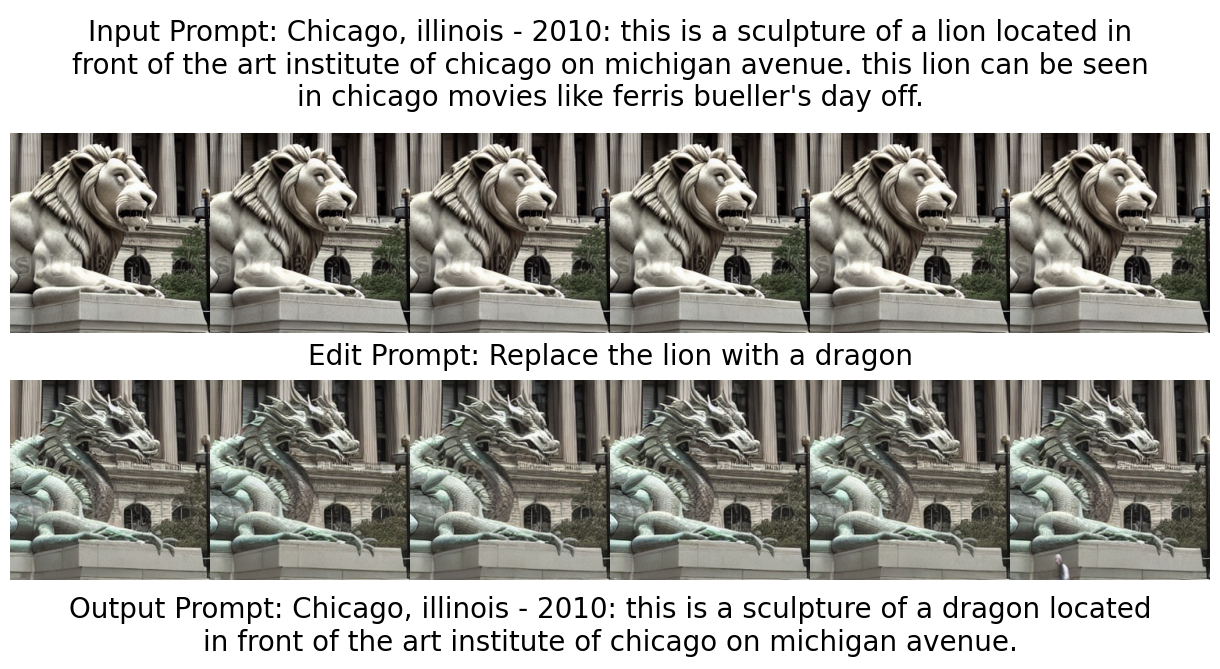}
    \includegraphics[width=\linewidth]{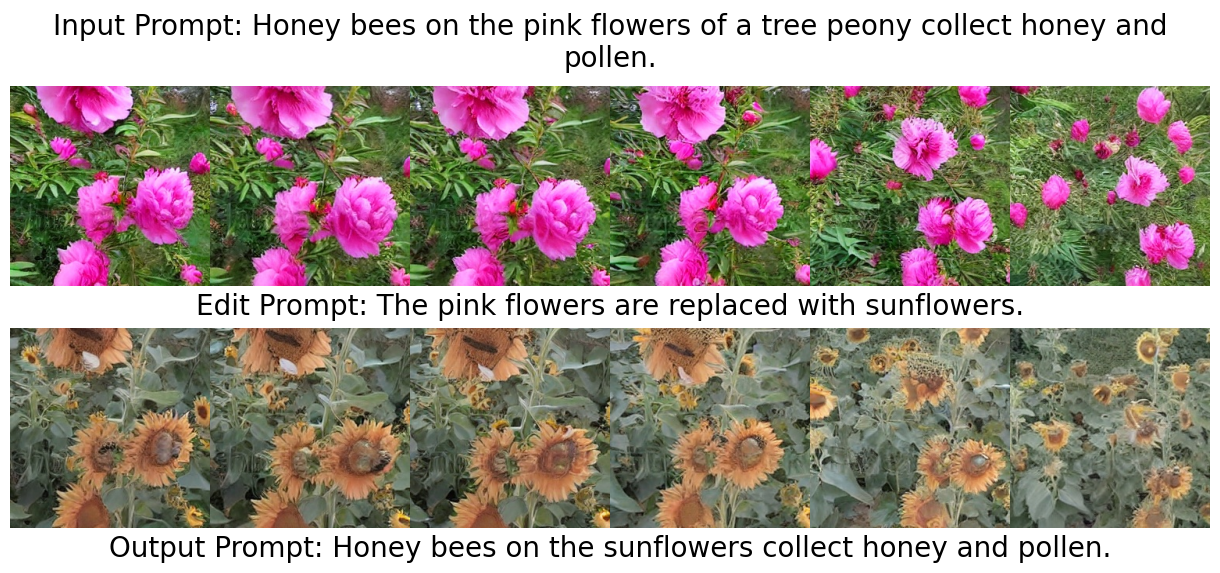}
    \includegraphics[width=\linewidth]{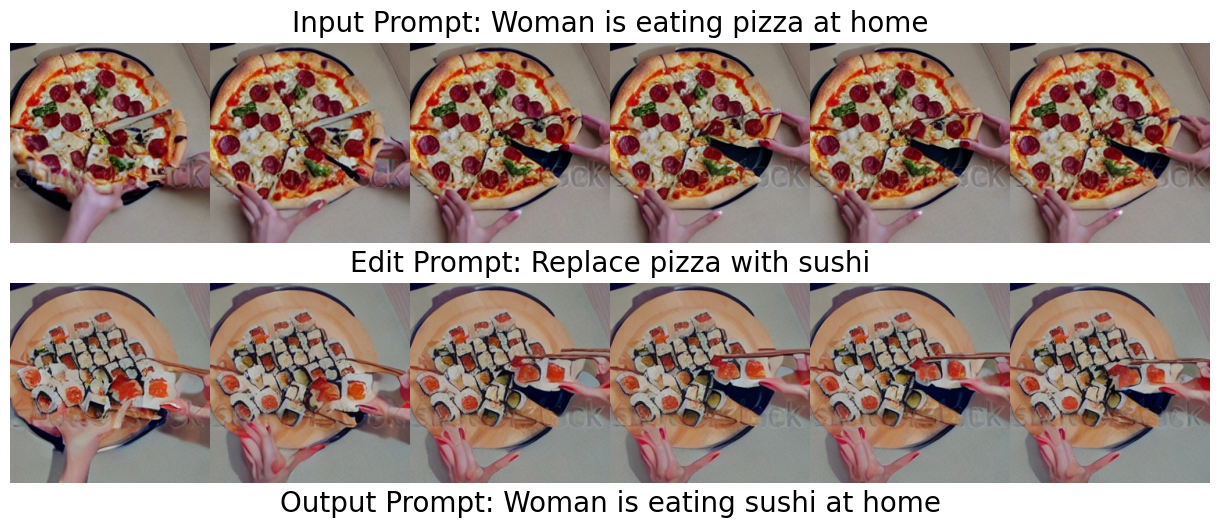}
    \caption{Visualization of synthetic samples produced by the data generation pipeline in \Cref{sec.synthetic_data_pipe}. Each sample contains two generated videos: the upper one represents the input prompt, while the lower one depicts the output prompt. The actual videos comprise 16 frames while 6 subsampled frames are displayed here for brevity.}
    \label{fig.synthetic_vid2}
\end{figure}

In \Cref{sec.synthetic_data_pipe}, we introduce a pipeline designed to produce synthetic paired videos. Further visual demonstrations of these results can be found in \Cref{fig.synthetic_vid1,fig.synthetic_vid2}. The generated paired videos maintain a similar structure, and the editing can showcase the edit prompt.

\section{Long Video Sampling Detail Illustration and Examples}\label{sec.long_vid_editing_appendix}
\begin{figure}
    \centering
    \includegraphics[width=\linewidth]{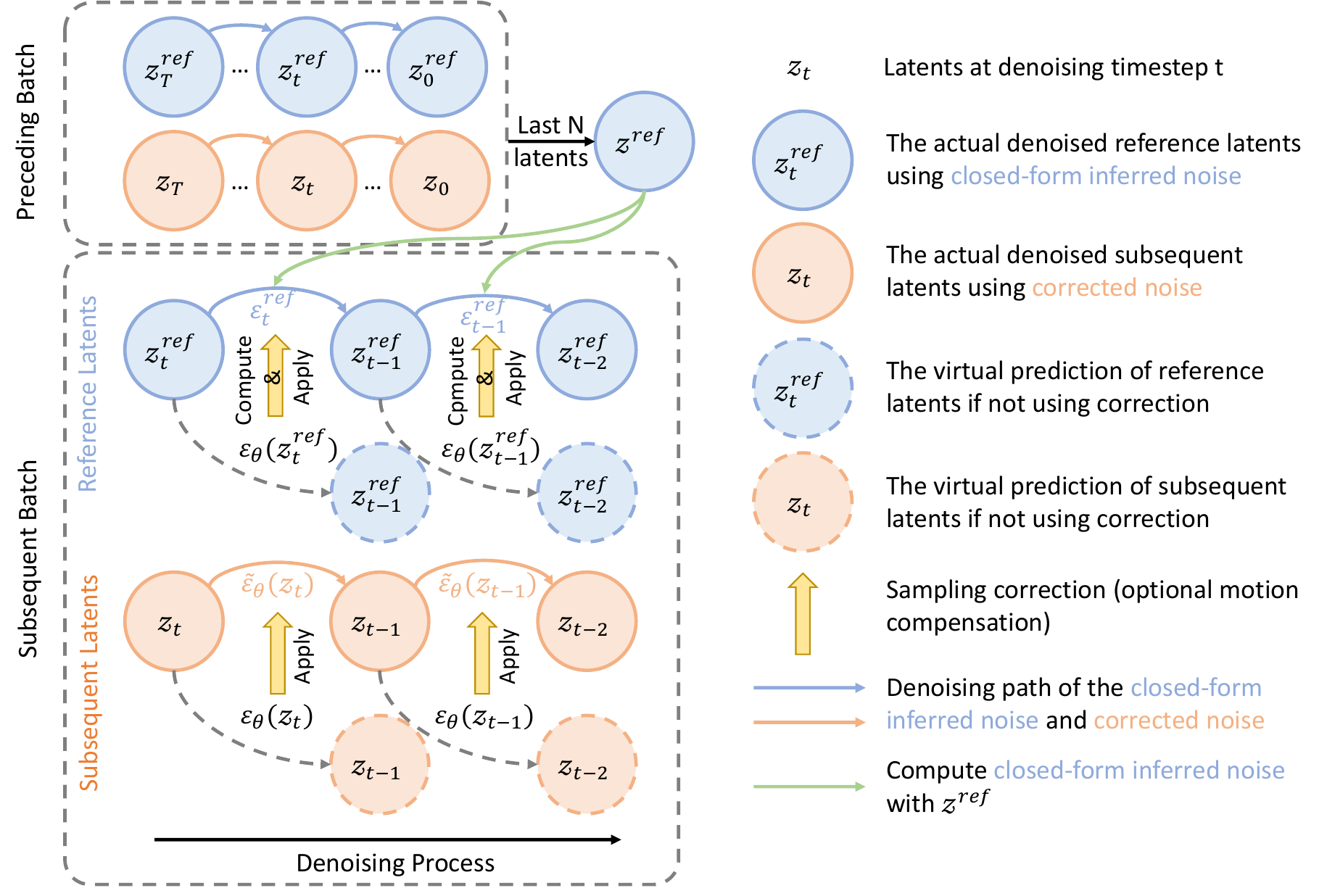}
    \caption{Schematic of Long Video Sampling Correction (LVSC) mechanism. The diagram illustrates the interaction between reference latents from a preceding batch and subsequent latents during the denoising process. The closed-form inferred noise (\Cref{eq.1}) is computed for the $N$ reference latents (shown in blue), which then guides the correction of the actual denoised subsequent latents (shown in orange).}
    \label{fig.lvsc_detail}
\end{figure}

In Section \ref{sec.long_vid_editing}, we addressed the challenge of ensuring consistency across different batches when sampling long videos, particularly when each batch is processed separately. This section introduces an illustrative explanation of the Long Video Sampling Correction (LVSC) method in Figure \ref{fig.lvsc_detail}. In LVSC, there are $N$ overlapping reference frames between two consecutive batches, but this method transcends a basic sliding window technique. Instead of allowing the editing model unrestricted freedom in transferring edits, the reference frames guide the appearance of subsequent frames during the editing process.

\begin{figure}
    \centering
    Edit Prompt: Change the watermelon to hamburger.
    \includegraphics[width=\linewidth]{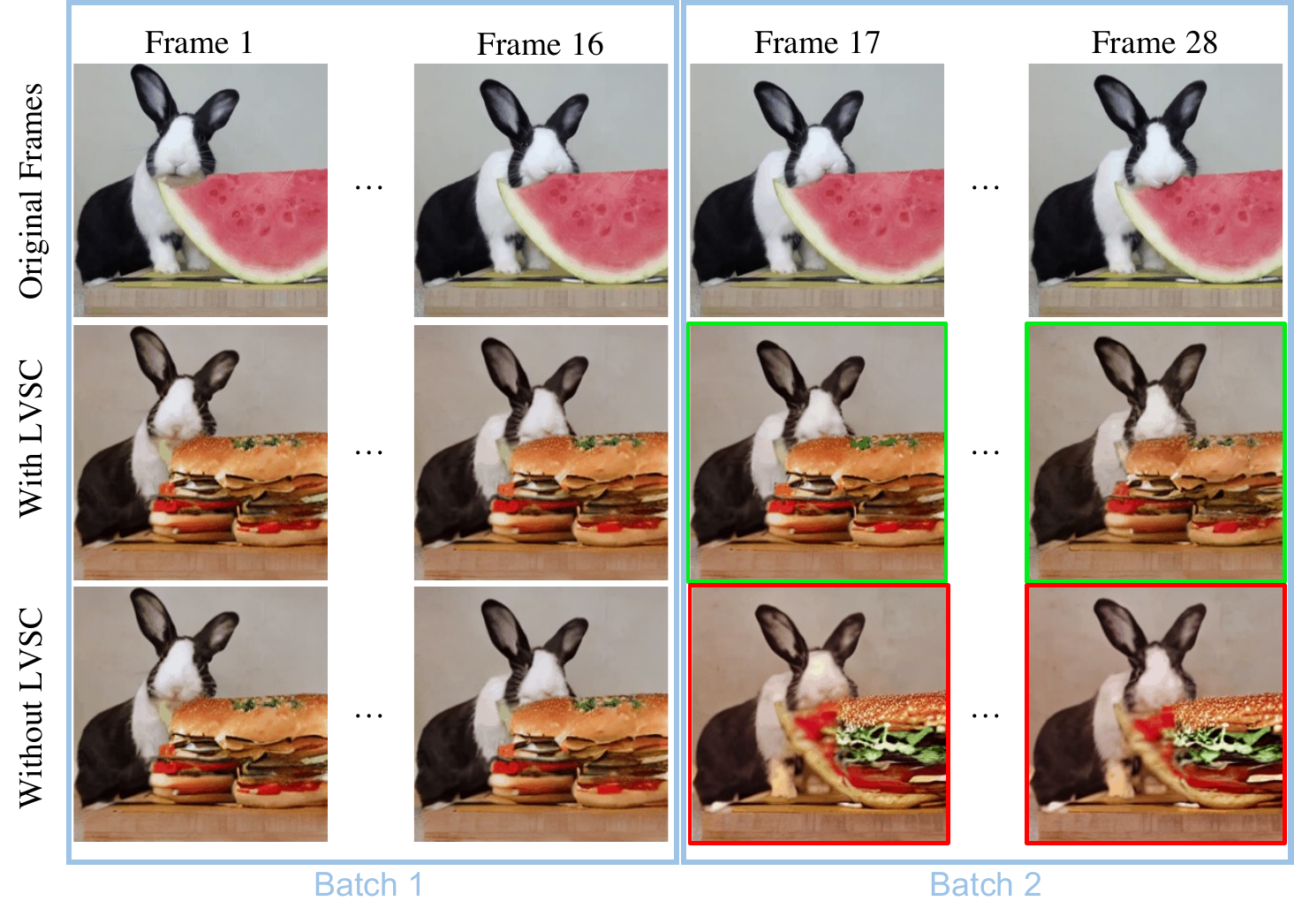}
    \caption{In the process of video sampling, we utilize a batch size of 16. The 17th to 28th frames in the video are processed in the second batch, and they reference the last four sampled frames from the preceding batch. By employing the Long Video Sampling Correction (LVSC), we can ensure the content consistency between sampled frames across different batches (green boxes in the figure). In contrast, sampling two batches separately may lead to inconsistencies at the boundaries where the batches change  (rex boxes in the figure).}
    \label{fig.long_vid_sampling}
\end{figure}

Additionally, we present a visual comparison to demonstrate the impact of LVSC. Figure \ref{fig.long_vid_sampling} contrasts the results with and without LVSC implementation. This comparison clearly shows the lack of continuity in the sampled frames between batches when LVSC is not applied. In contrast, employing LVSC achieves noticeable consistency, aligning the first frame of a subsequent batch with the last frame of the previous batch.

\section{Effect of Motion Compensation in Long Video Sampling Correction}\label{sec.motion_compensation}
\begin{figure}
    \centering
    Edit Prompt: Make it Van Gogh Starry Night style.
    \includegraphics[width=\linewidth]{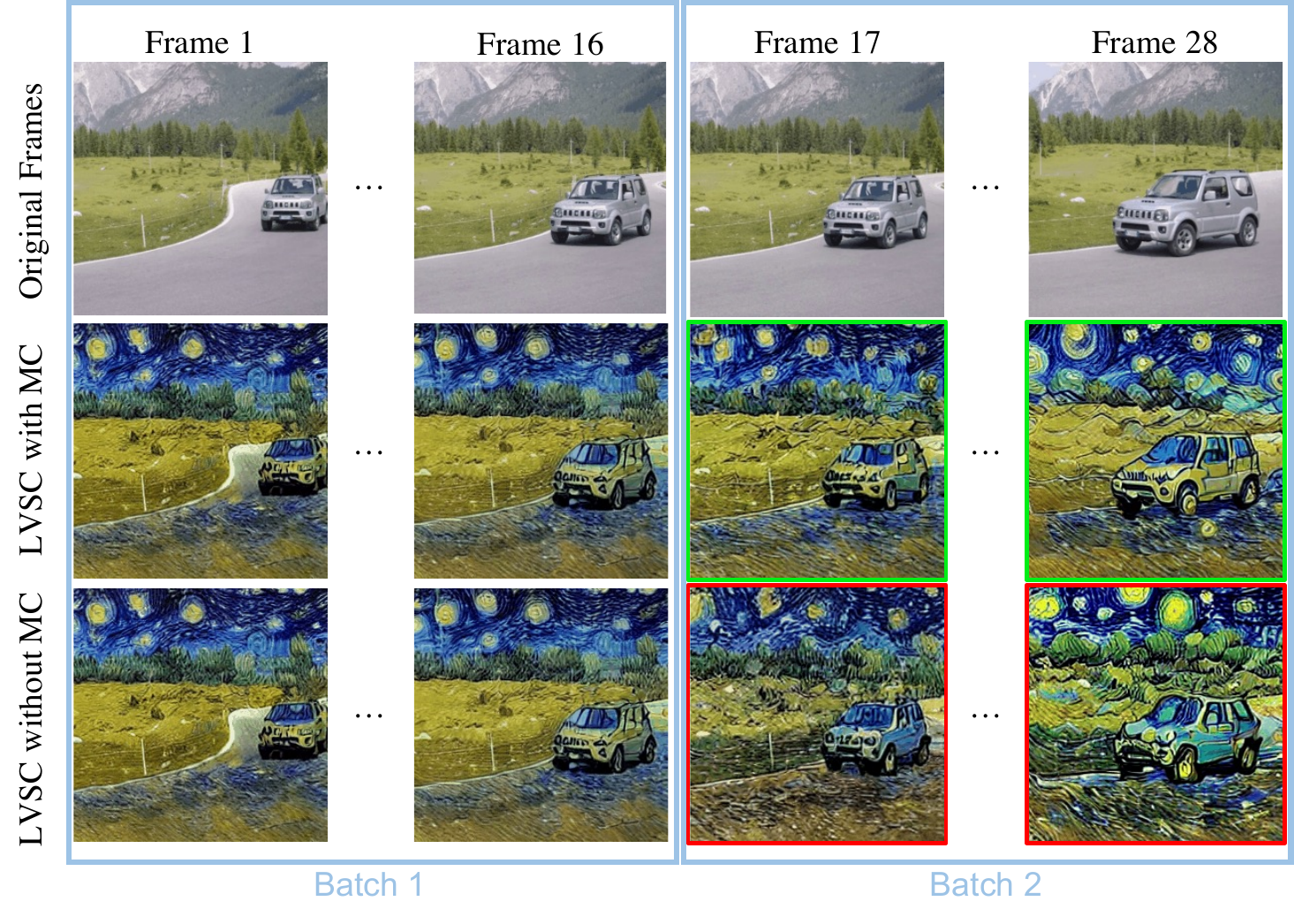}
    \caption{When the video exhibits extensive camera motion, the LVSC model without motion compensation (LVSC without MC) suffers significant quality degradation, particularly in frames from subsequent batches (red boxes in the figure). Incorporating motion compensation (LVSC with MC) substantially improves frame consistency with the reference and maintains high transfer quality, even when the camera view changes dramatically in later batches (green boxes in the figure). }
    \label{fig.motion_compensation}
\end{figure}

In our experiments, we observed that the presence of holistic camera motion could degrade the transfer results of subsequent batches processed by the LVSC model (see red boxes in \Cref{fig.motion_compensation}). This degradation arises because the same regions across different frames require consistent score corrections. In other words, score corrections should ``travel'' with the regions affected by camera movement, ensuring that identical corrections are applied to the same areas regardless of motion. To address this issue, we introduce a motion compensation strategy. Specifically, we first employ the RAFT flow estimator~\cite{teed2020raft} to compute the optical flow between each reference frame and the remaining frames in subsequent batches. The original equation for LVSC (\Cref{eq.lvsc}) is then modified as follows:
\begin{align}
\tilde{\varepsilon_\theta}(z_t)[:, m] &= \varepsilon_\theta(z_t)[:, m] + (\frac{1}{N} \sum_{i=1}^N o(\varepsilon_t^{ref}[:, i]), i \rightarrow m) - \varepsilon_\theta(z_t)[:, m]
\end{align}

Here, $m=[1, 2, ..., M]$ represents the indices of frames in the subsequent batches, which contain a total of $M$ frames, $o(\cdot, i \rightarrow m)$ is a warping function that uses optical flow to align the $i$-th reference frame with the $m$-th frame in the subsequent batch. This modification ensures better transfer quality by making the score corrections to be aware of camera movement. The green boxes in \Cref{fig.motion_compensation} reveal that applying motion compensation significantly enhances content consistency and overall quality. This improvement is particularly noticeable in the last few frames of the subsequent batches.

\section{Effect of Sampling Hyperparameters and Picking Criteria}\label{sec.pick_criteria}
\begin{figure}
    \centering
    Edit Prompt: Make the rabbit robotic.
    \includegraphics[width=\linewidth]{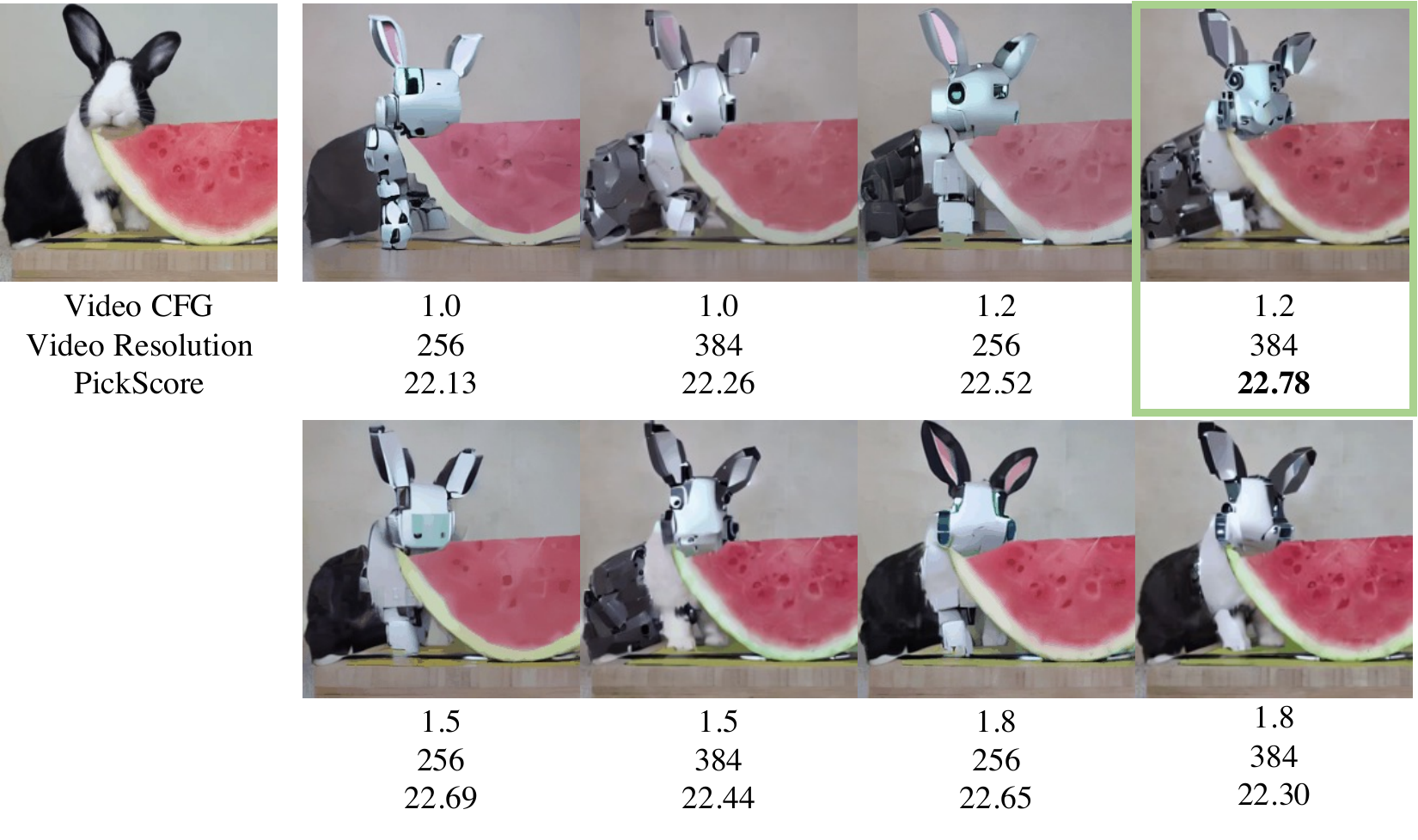}
    \caption{During sampling, the interplay between the scale of image classifier-free guidance (CFG) and image size can significantly influence the transfer result. In our evaluation, we sample videos by employing various combinations of CFG scales and resolutions, utilizing the average PickScore across all frames as the selection metric. The video with the highest PickScore is designated as the final transfer result. As indicated by the green box in the figure, the chosen video and its corresponding hyperparameters bear the highest PickScore, thereby making it the final selection.}
    \label{fig.pickscore_sample}
\end{figure}

In our experiments, we observed that the video CFG and resolution have a greater impact on the generated video than the text CFG. To streamline the parameter search process, we thus focus solely on picking the video CFG and resolution, effectively reducing the search space. Detailed visual effects of the hyperparameter choices are provided in \Cref{fig.pickscore_sample}. We opt for PickScore~\cite{kirstain2023pick} as our automated selection criterion. Our choice is motivated by the fact that PickScore aligns more closely with human perception compared to the CLIP score, as indicated in \cite{kirstain2023pick}.

\section{Qualitative Comparisons}\label{sec.qualitative_comparison}
\begin{figure}
    \centering
    
    \includegraphics[width=\linewidth]{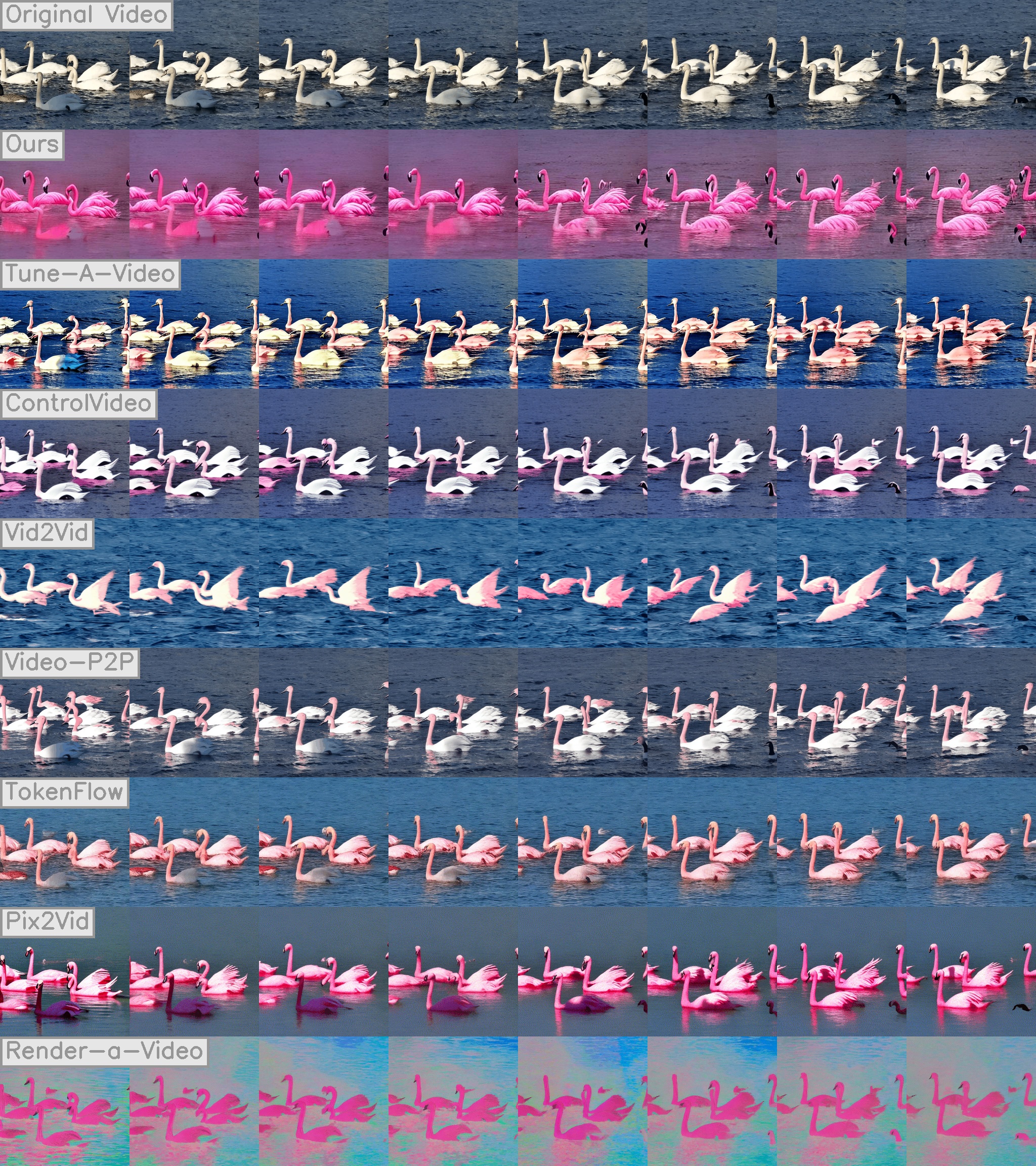}
    \caption{\textbf{Swans} gliding over a lake. $\rightarrow$ \textbf{Pink flamengos} gliding over a lake.}\label{fig.additional_qua1}
\end{figure}

\begin{figure}
    \centering
    \includegraphics[width=\linewidth]{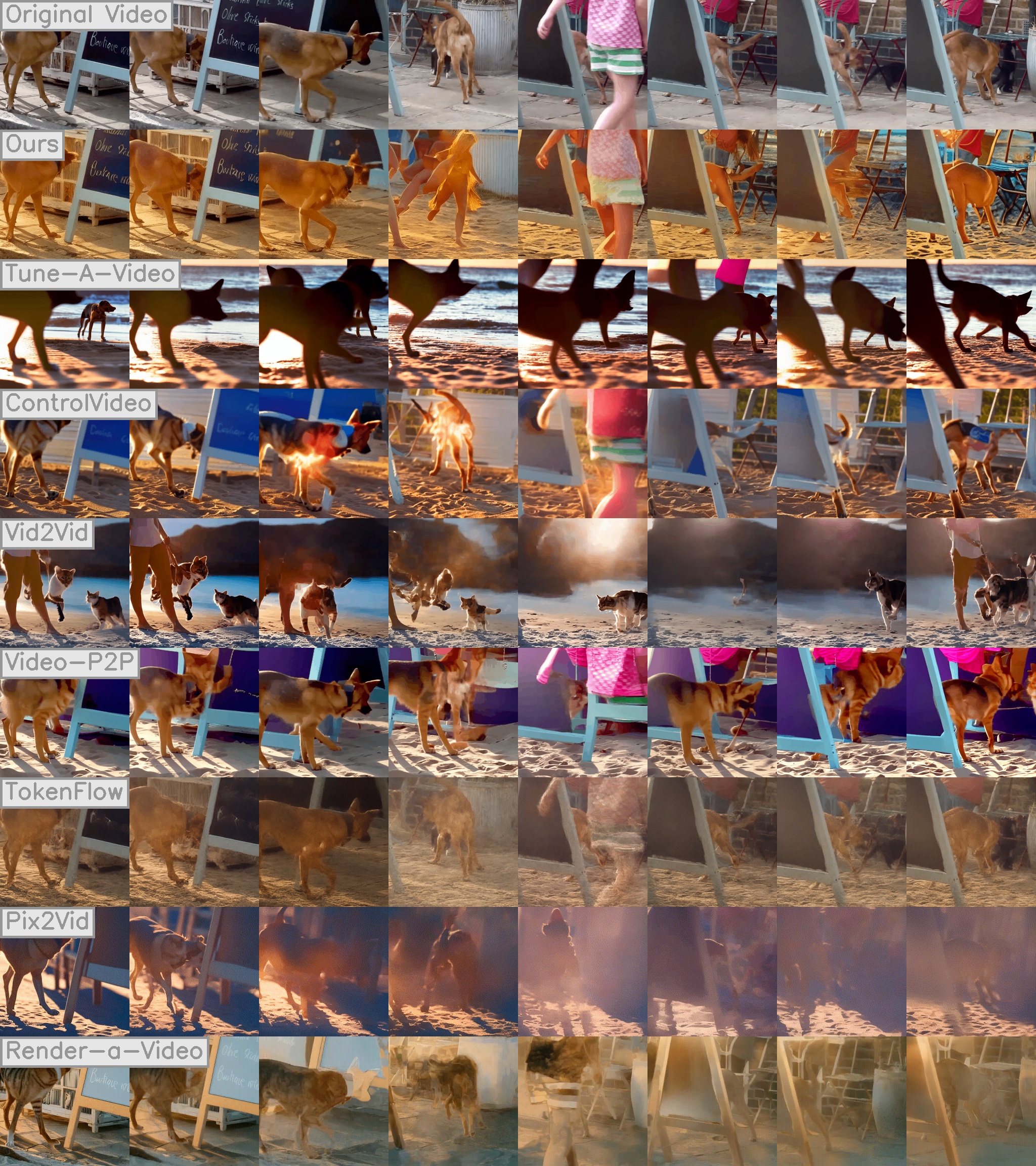}
    \caption{A cat and a dog playing on the street while a girl walks around them. \\ $\rightarrow$ A cat and a dog playing on the beach while a girl walks around them, \textbf{golden hour lighting}.}\label{fig.additional_qua2}
\end{figure}

\begin{figure}
    \centering
    \includegraphics[width=\linewidth]{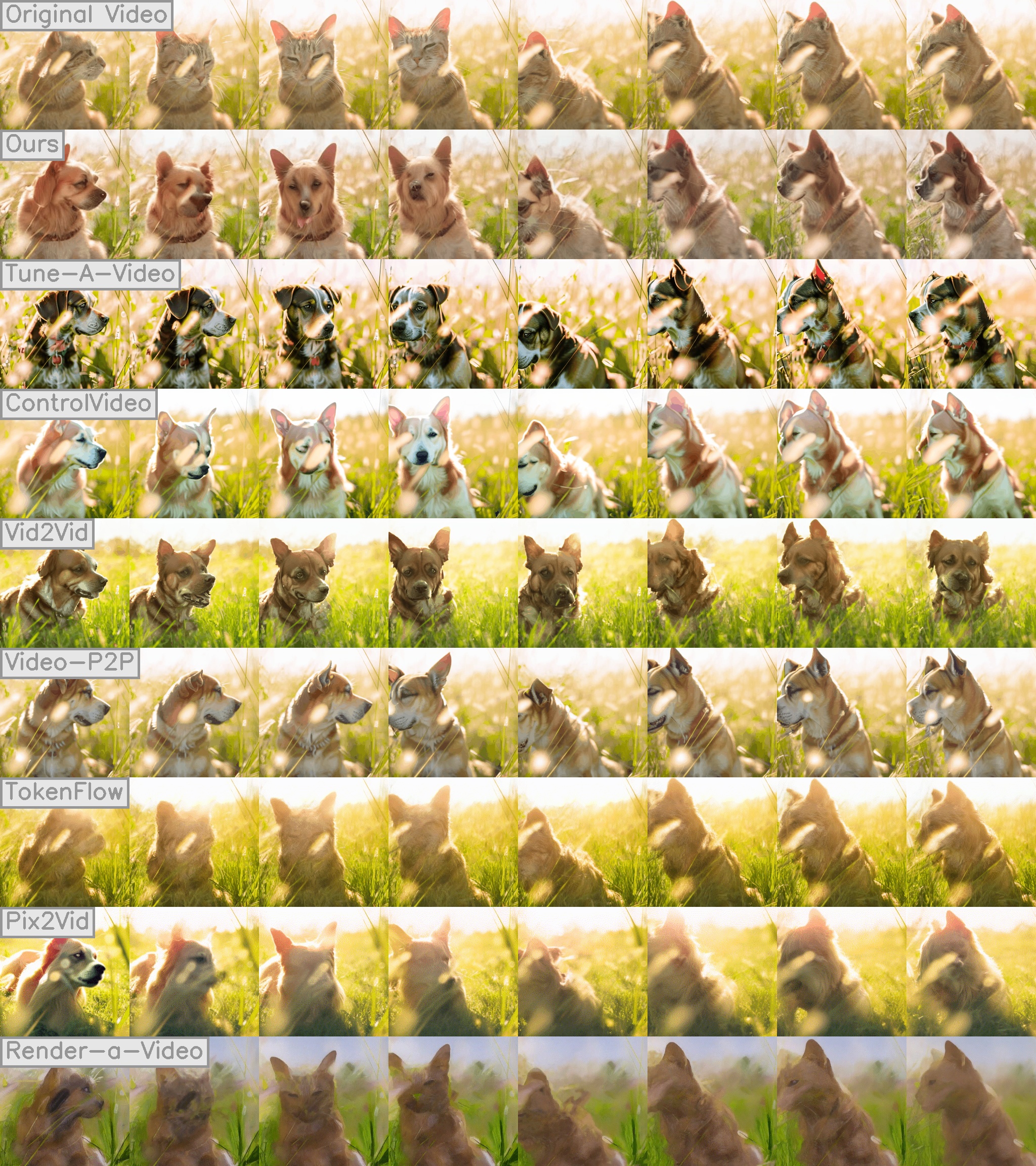}
    \caption{A \textbf{cat} in the grass in the sun. $\rightarrow$ A \textbf{dog} in the grass in the sun.}\label{fig.additional_qua3}
\end{figure}

\begin{figure}
    \centering
    \includegraphics[width=\linewidth]{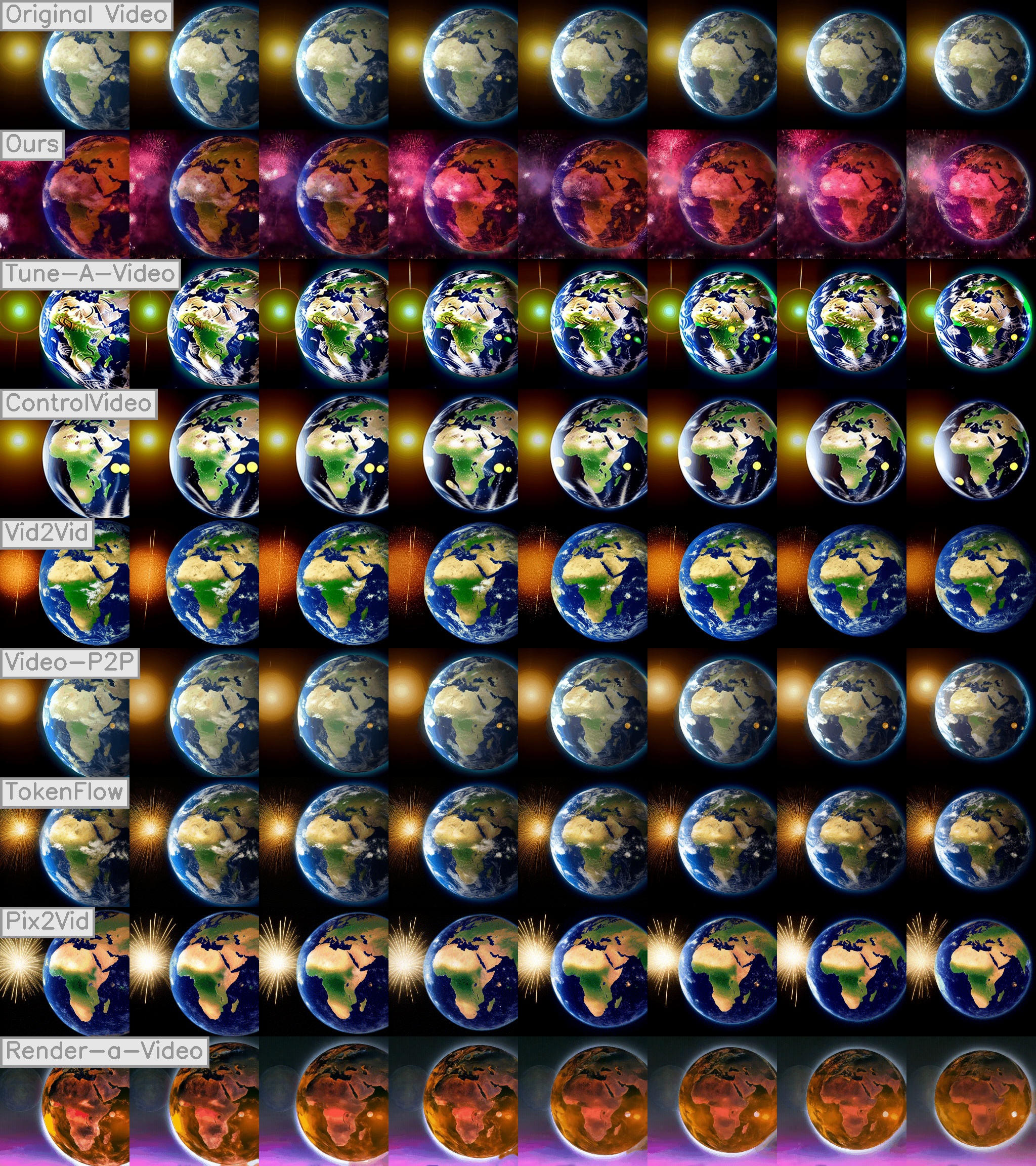}
    \caption{Full view of the Earth as it moves slowly \textbf{toward the sun}. \\ $\rightarrow$ Full view of the Earth as it moves slowly \textbf{through a fireworks display.}}\label{fig.additional_qua4}
\end{figure}

\begin{figure}
    \centering
    \includegraphics[width=\linewidth]{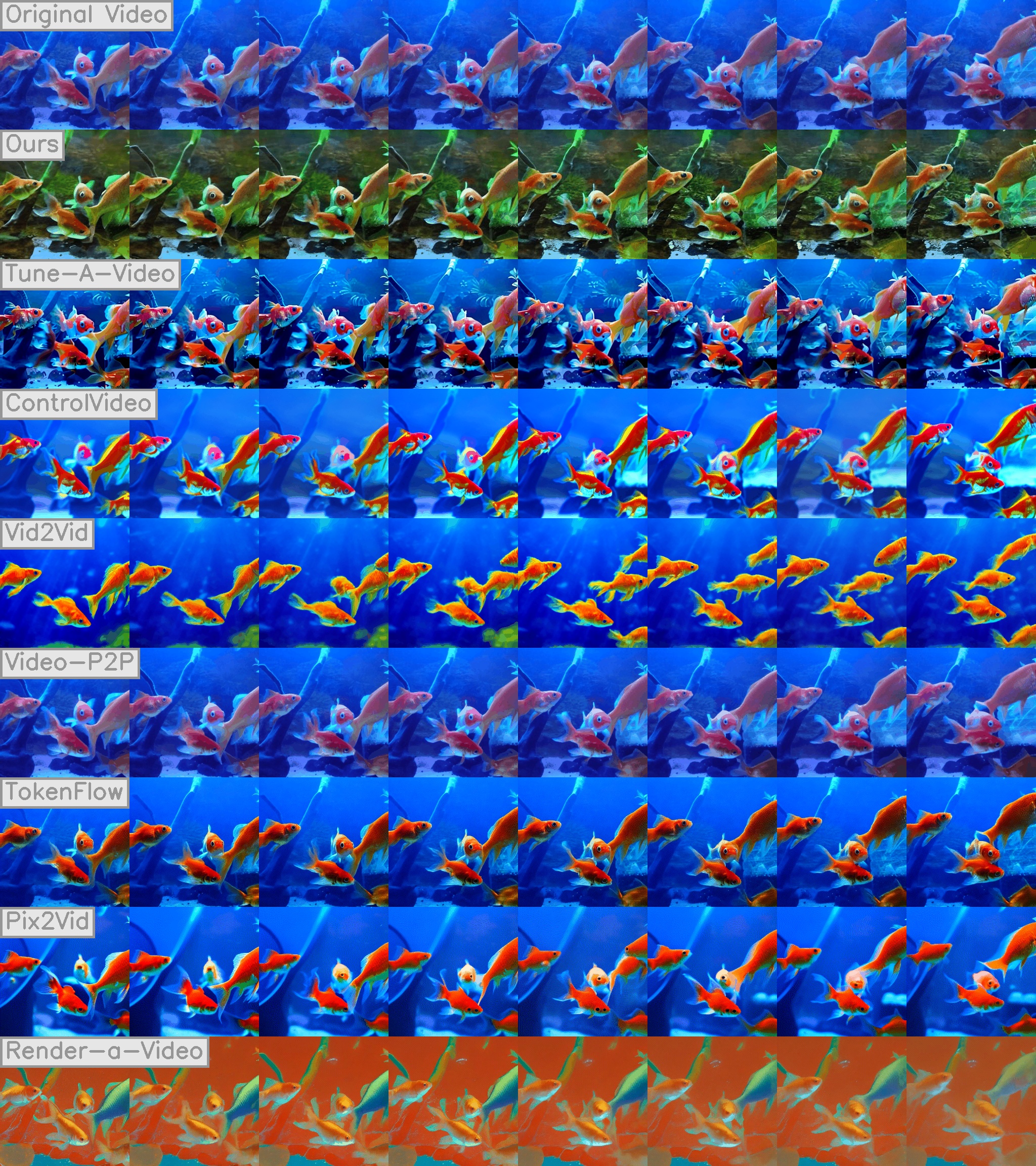}
    \caption{Several goldfish swim in a \textbf{tank}. \\ $\rightarrow$ Several goldfish swim in a \textbf{pond}.}
    \label{fig.additional_qua5}
    
\end{figure}

\begin{figure}
    \centering
    \includegraphics[width=\linewidth]{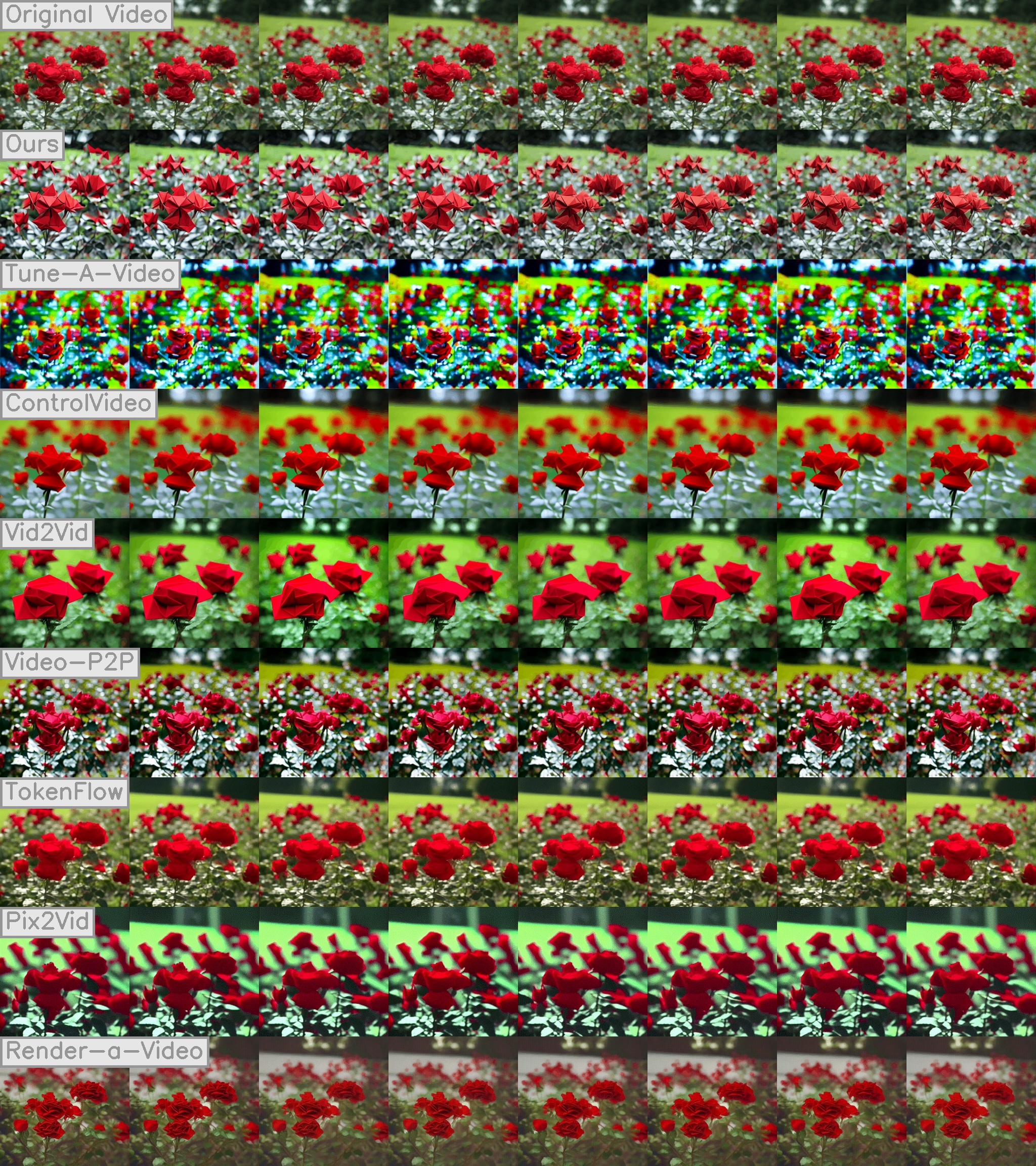}
    \caption{A static shot of red roses in sunlight, gently swaying in the breeze. \\ $\rightarrow$ A static shot of red roses in sunlight, gently swaying in the breeze, \textbf{origami style}.}
    \label{fig.additional_qua6}
\end{figure}

\begin{figure}
    \centering
    \includegraphics[width=\linewidth]{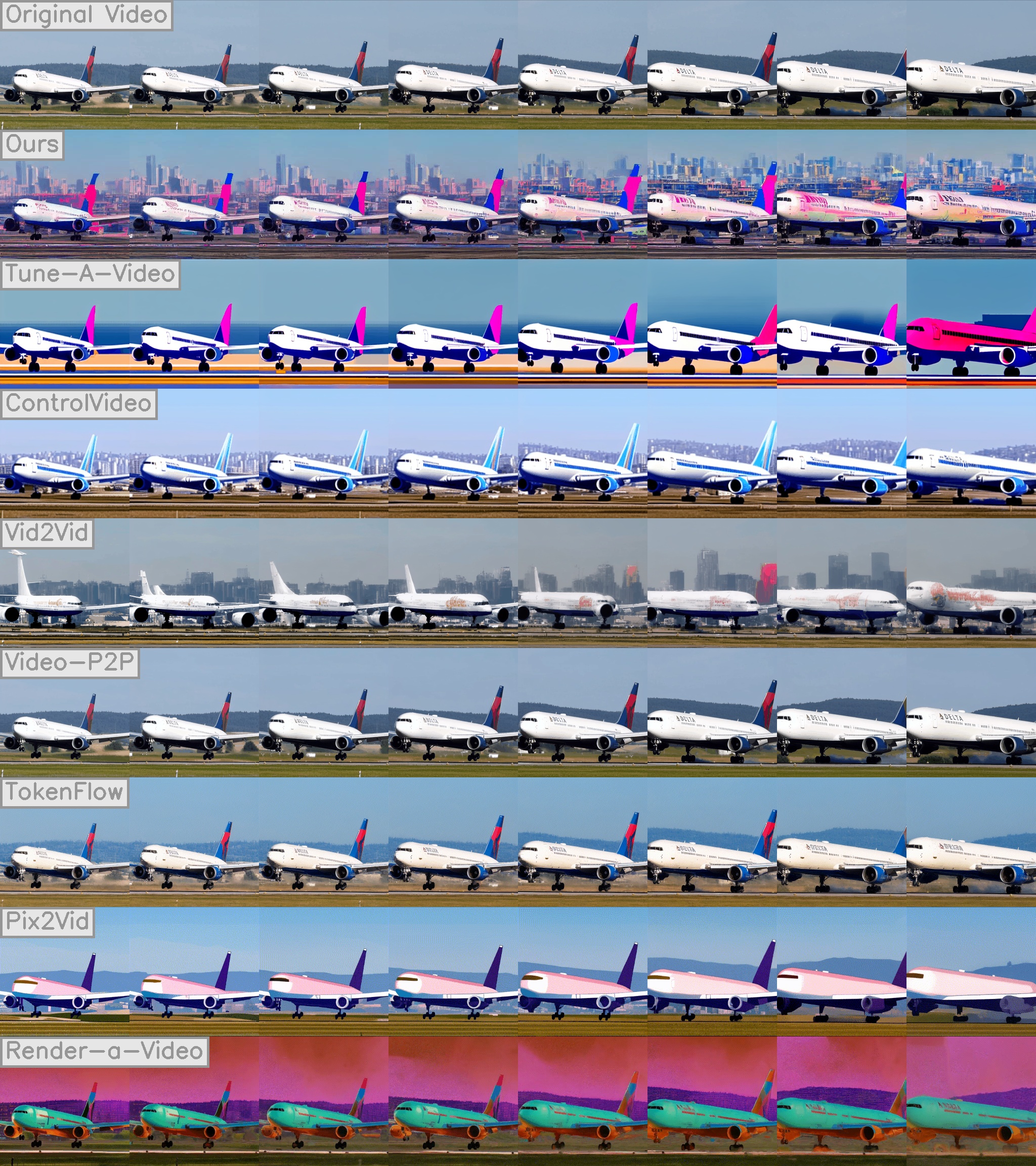}
    \caption{A Delta Airlines aircraft descends onto the runaway during a cloudless morning. \\ $\rightarrow$ A \textbf{brightly colored cyberpunk aircraft} descends onto the runway during a cloudless morning, \textbf{with a bustling cityscape in the background}.}
    \label{fig.additional_qua7}
\end{figure}

\begin{figure}
    \centering
    \includegraphics[width=\linewidth]{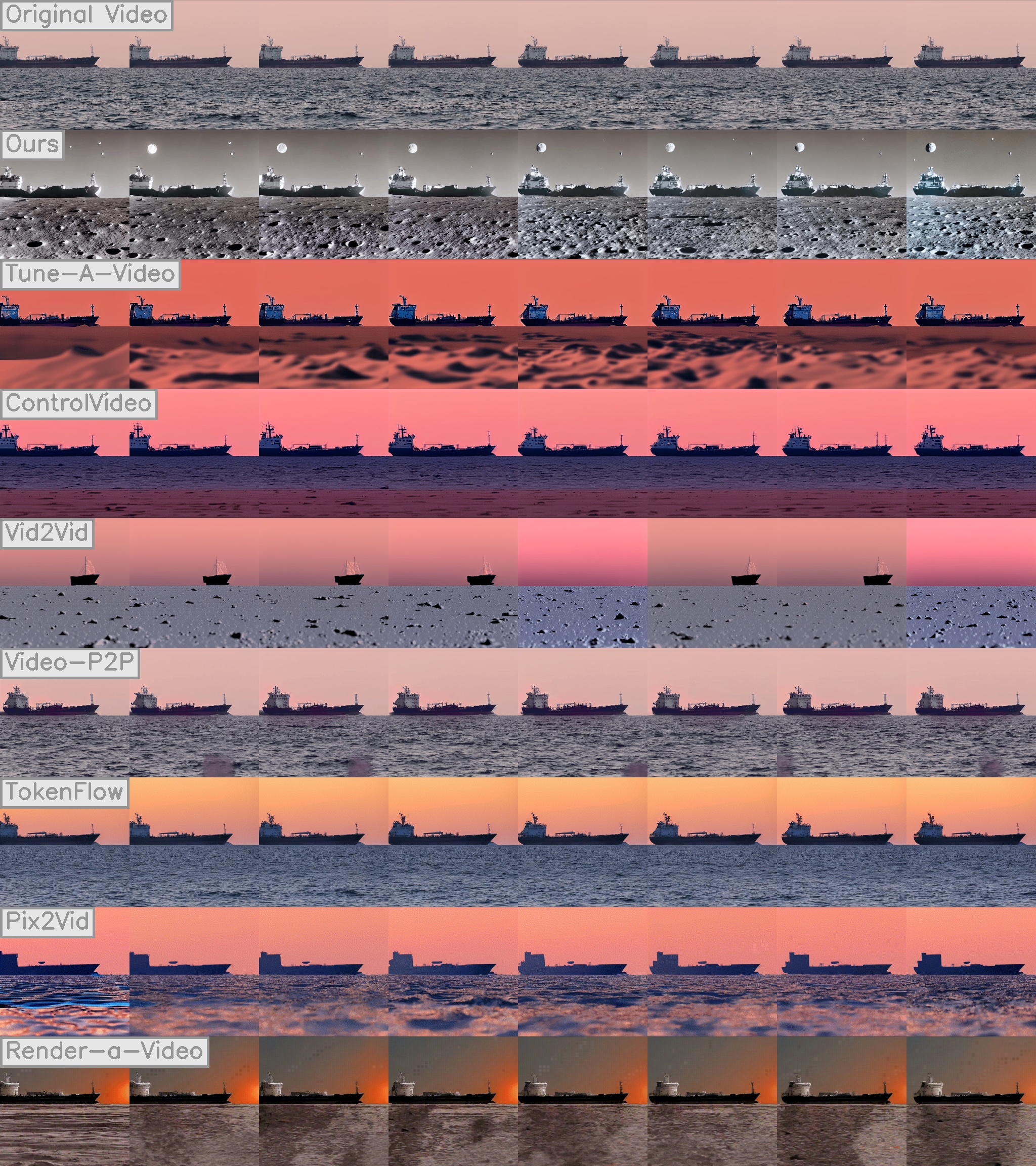}
    \caption{A ship sails on the \textbf{sea}. \\ $\rightarrow$ A ship sails on the \textbf{lunar surface}.}
    \label{fig.additional_qua8}
\end{figure}

\begin{figure}
    \centering
    \includegraphics[width=\linewidth]{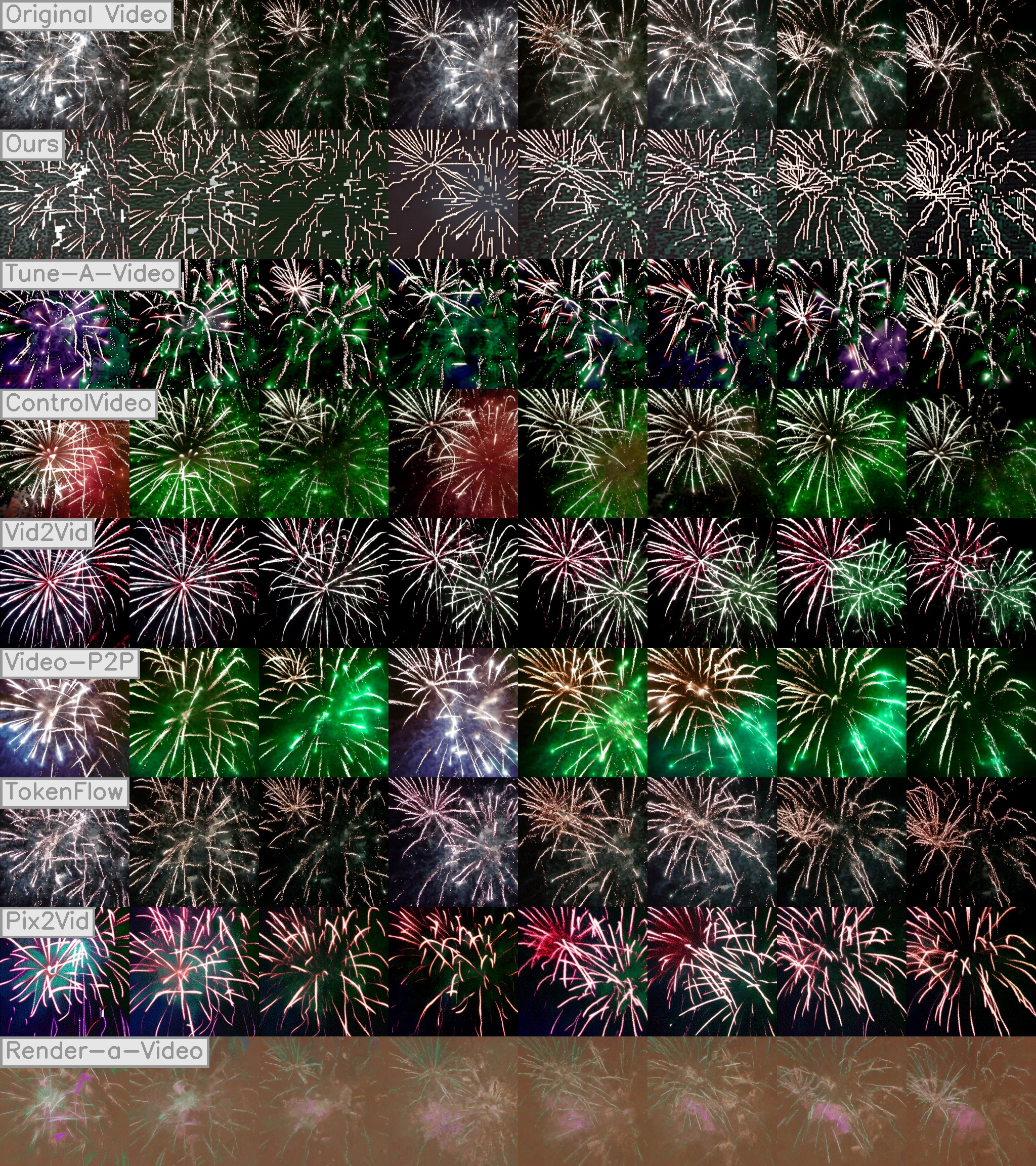}
    \caption{A colourful fireworks display in the night sky. \\ $\rightarrow$ A colourful fireworks display in the night sky, \textbf{8-bit pixelated}.}
    \label{fig.additional_qua9}
\end{figure}

We provide qualitative comparisons with baselines in ~\Cref{fig.additional_qua1,fig.additional_qua2,fig.additional_qua3,fig.additional_qua4,fig.additional_qua5,fig.additional_qua6,fig.additional_qua7,fig.additional_qua8,fig.additional_qua9}

\section{Failure Cases}
Our model, \ours, occasionally encounters challenges in video transfer, particularly when the object of interest is difficult to detect. Such difficulties arise when the object is positioned close to the video's edge, is unusually small or large, or is partially obscured. These scenarios can lead to the object either vanishing in the edited video or exhibiting inconsistent appearances.

\begin{figure}
    \centering
    Prompt: Change background to tropical river.
    \includegraphics[width=\linewidth]{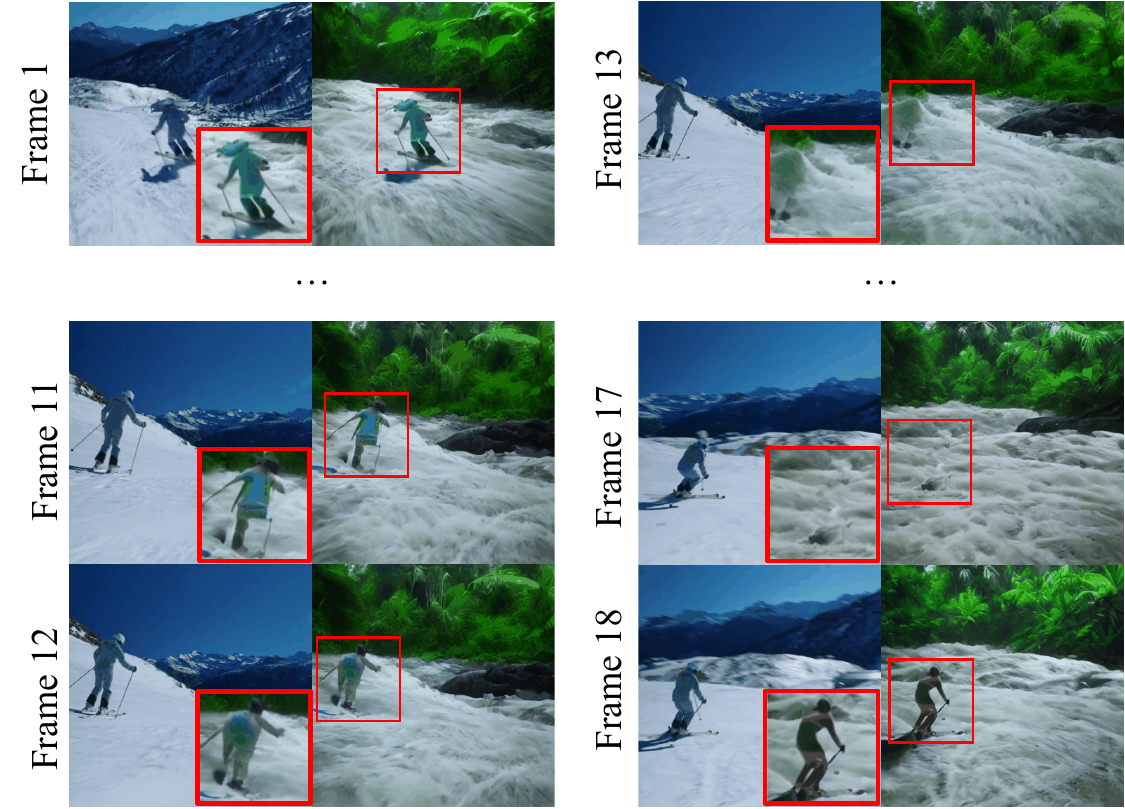}
    \caption{Illustration of a failure case where \ours\ struggles to maintain the structure of a person located near the video's edge and occupying a small area, resulting in the disappearance of the person from the transferred video.}
    \label{fig.failure1}
\end{figure}

For instance, as depicted in \Cref{fig.failure1}, when a person is situated near the frame's border and occupies a small area, \ours\ struggles to maintain the person's structural integrity, resulting in their disappearance from the transferred video. However, when the person moves away from the border, \ours\ can recognize them again, but with a notably altered appearance.

\begin{figure}
    \centering
    Prompt: Make trucks made of wood (wooden style truck).
    \includegraphics[width=\linewidth]{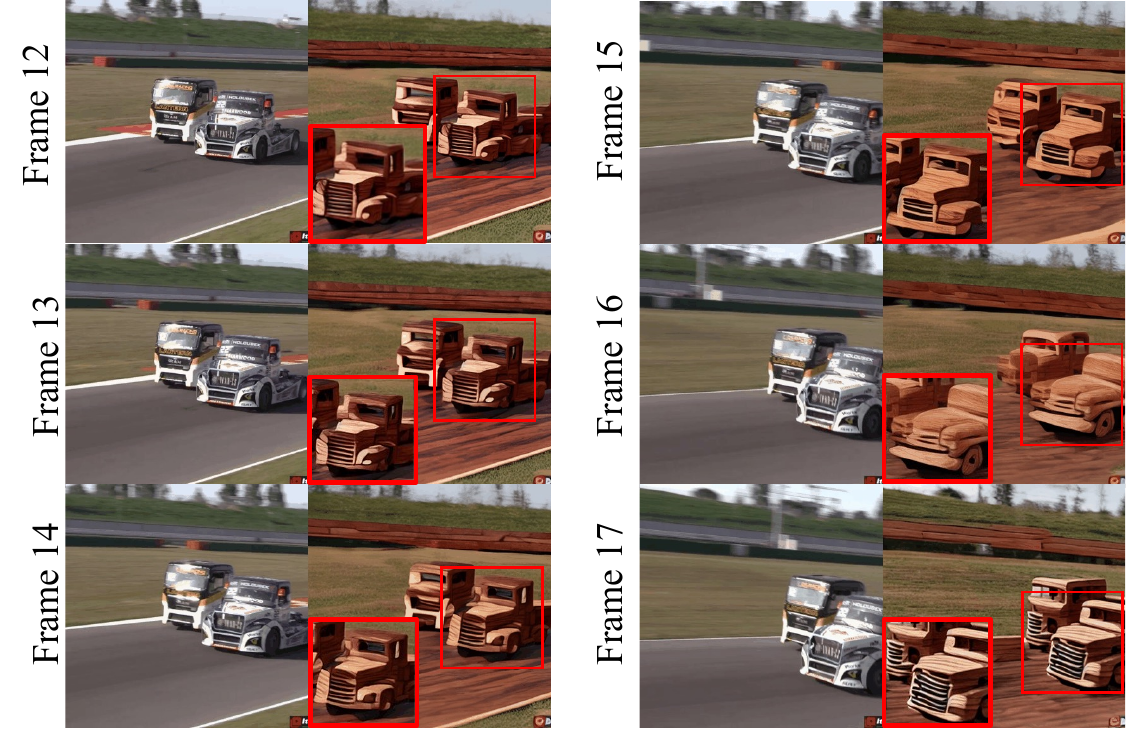}
    \caption{Demonstration of a failure scenario in \ours\ where a partially visible truck is misidentified as a car's front, leading to inconsistent transfer results in the edited video.}
    \label{fig.failure2}
\end{figure}

\Cref{fig.failure2} presents another scenario where the object, in this case, the front part of a truck, is only partially visible in certain frames. While \ours\ can accurately transfer the image when the truck is fully visible, it misinterprets the partially visible truck as a car's front, leading to inconsistent results in the transferred video.

In summary, when object detection is hindered due to size, positioning, or occlusion, \ours\ may not deliver satisfactory transfer results.

